\documentclass[10pt,twocolumn,letterpaper]{article}

\usepackage[pagenumbers]{wacv}

\usepackage{multirow}
\usepackage{makecell}
\usepackage{wrapfig}
\usepackage{float}
\usepackage{nicefrac}
\usepackage{microtype}
\usepackage{dblfloatfix}
\usepackage{placeins}
\usepackage{colortbl}
\usepackage{multirow}
\usepackage{wrapfig}
\usepackage{caption}                 %
\newsavebox{\tabOne}\newsavebox{\tabGen}\newsavebox{\tabQual}


\definecolor{wacvblue}{rgb}{0.21,0.49,0.74}
\usepackage[pagebackref,breaklinks,colorlinks,allcolors=wacvblue]{hyperref}

\def\wacvPaperID{615} %
\def\confName{WACV}
\def\confYear{2027}

\title{ComposeAnything: Composite Object Priors for Text-to-Image Generation}

\author{
Zeeshan Khan \qquad Shizhe Chen \qquad Cordelia Schmid\\
Inria, \'Ecole normale sup\'erieure, CNRS, PSL Research University\\
{\small \url{https://zeeshank95.github.io/composeanything/ca.html}}
}

\begin{document}
\twocolumn[{%
  \renewcommand\twocolumn[1][]{#1}%
  \maketitle
}]

\begin{abstract}
\label{sec:abs}

Layout-based compositional methods improve text-to-image generation by specifying where objects should appear, but most rely on sparse 2D constraints with limited appearance, occlusion, and depth-aware cues.
We show that compositional generation can be substantially improved by replacing purely geometric guidance with an editable visual prior for diffusion initialization.
We introduce ComposeAnything, an inference-time framework built around a 2.5D Composite Object Prior, an editable RGB image that specifies object appearance, layout, scale, and depth-aware ordering.
We transfer this prior through Prior-Guided Diffusion, which combines Object Prior Reinforcement (OPR) to preserve object structure and appearance with Spatial-Controlled Denoising (SCD) to improve region-text alignment and reduce semantic leakage.
Early guidance preserves the prior structure, after which standard denoising refines the image coherently.
Extensive evaluations on T2I-CompBench, NSR-1K, and GenEval show strong gains in spatial reasoning, 3D-spatial relations, counting, and surreal compositions while maintaining visual quality and efficiency.

\end{abstract}

\section{Introduction}
\label{sec:intro}

Despite rapid progress in text-to-image (T2I) generation~\cite{SDXL,SD3,FLUX}, modern models still struggle with complex compositional prompts~\cite{huang2023t2icompbench}. When prompts require exact counts, unusual combinations, or explicit spatial and depth-aware relations, models often omit objects, duplicate instances, misbind attributes, or collapse the intended arrangement.

Existing compositional-control methods often improve prompt adherence through additional training~\cite{gligen,creatilayout,instancegen} or reward optimization~\cite{ddpo,dpok,flowgrpo}. These strategies can introduce task- or reward-specific distribution shifts, sacrificing visual quality for compositional accuracy \cite{flowgrpo}. Inference-time approaches offer a more flexible alternative by preserving the pretrained generator while steering its sampling process, but current methods remain limited. Layout-guided methods provide spatial constraints, yet most rely on limited geometric controls like 2D boxes, masks, or regions~\cite{layoutgpt,RPG,boundedattention,instancediffusion} that mainly control object positions while leaving appearance, occlusion, and 3D spatial relations ambiguous. Noise-search and inference-scaling methods exploit diffusion's sensitivity to initialization~\cite{ma2025inference,inito,ReflectDIT,CARINOX}, but require costly sampling or optimization and offer limited explicit control. As a result, layout-guided methods leave key visual and depth-aware structure to the model's learned prior, while noise-search methods seek favorable initializations without explicit compositional guidance. These limitations make complex or uncommon prompts difficult to satisfy reliably.

We address these limitations by turning structured layout guidance into a visual diffusion initialization. We introduce \textbf{ComposeAnything}, an inference-only framework that converts a 2.5D layout plan into an editable RGB \emph{Composite Object Prior} and injects it as a guided initialization during denoising. Rather than using boxes only to constrain attention or searching over random noises, the prior gives diffusion an explicit starting point that encodes object identity, appearance, count, scale, placement, and depth-aware ordering. Given a prompt, a structured planner predicts object descriptions, bounding boxes, and relative depth values; object instances are then generated, segmented, and composed into the prior.

The central challenge is transferring this structured initialization into a coherent final image while preserving the generative flexibility of the pretrained model. We propose \emph{Prior-Guided Diffusion}, which operates during the early structural stage of denoising. \emph{Object Prior Reinforcement (OPR)} anchors the denoising trajectory to the prior so that object identity, appearance, count, placement, and depth ordering are preserved. \emph{Spatial-Controlled Denoising (SCD)} strengthens the correspondence between object regions and their textual descriptions, reducing semantic leakage across regions. During these early steps, the prior anchors the specified objects while the background and global scene are synthesized around them. Afterward, the guidance is removed and standard denoising refines the full image coherently, preserving the base model's flexibility for lighting, natural object interactions, background, and global appearance.

We evaluate ComposeAnything on T2I-CompBench~\cite{huang2023t2icompbench}, NSR-1K~\cite{layoutgpt}, and GenEval~\cite{geneval}. Across these benchmarks, ComposeAnything improves 2D spatial accuracy, object counting, 3D-spatial relations, and complex surreal compositions while maintaining strong visual quality and computational efficiency. Further analyses validate the controllability and robustness of prior-guided diffusion, with consistent gains across multiple base models. The editable prior also enables simple manual refinements that further improve difficult compositions.

Our contributions are three-fold:
\begin{itemize}
\item We introduce an editable 2.5D Composite Object Prior that encodes object appearance, layout, scale, and depth as a visual initialization for diffusion.

\item We propose Prior-Guided Diffusion, combining Object Prior Reinforcement and Spatial-Controlled Denoising to transfer prior structure and appearance while preserving the generative flexibility of the pretrained model.

\item We demonstrate strong improvements across spatial, counting, 3D-spatial relations, and complex-composition benchmarks while maintaining visual quality.
\end{itemize}

\section{Related Works}
\label{sec:related_works}

\begin{figure*}[t]
    \centering
    \includegraphics[width=0.65\textwidth]{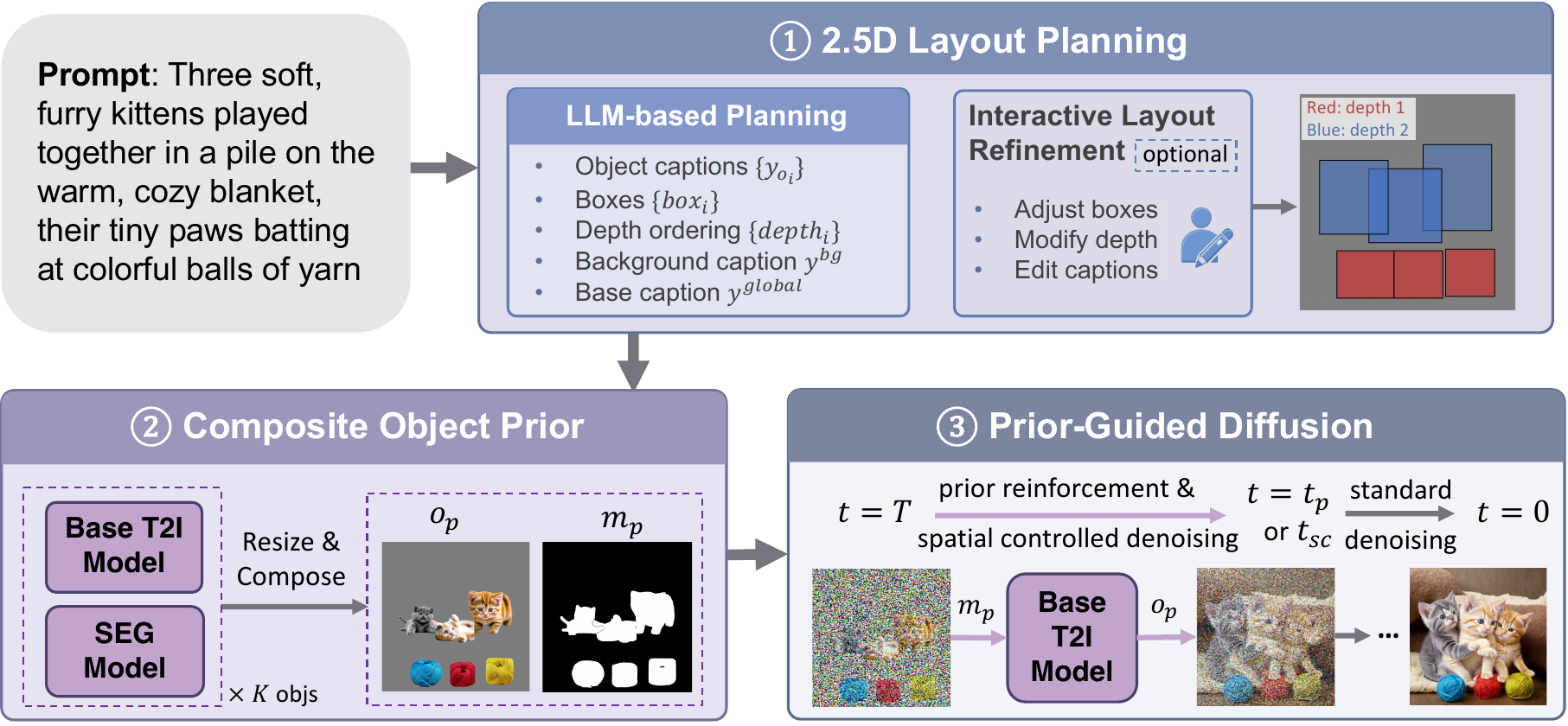}
    \caption{The \emph{ComposeAnything} framework, which enhances text-to-image diffusion models with layouts and composite object priors for complex compositional generation. 
    }
    \label{fig:framework}
    \vspace{-1em}
\end{figure*}

\noindent\textbf{Compositional generation.}
Compositional T2I aims to produce images faithful to complex textual descriptions~\cite{huang2023t2icompbench,2024realcompo,2024compositediff,Li_2024_CVPR,instancediffusion,layoutgpt,RPG,tokencompose,Couairon_2023_ICCV,lian2024llmgrounded}; while modern diffusion models~\cite{SDXL,SD3,FLUX} are strong generators, they struggle with novel multi-object arrangements, spatial relations, and counting.

\noindent\textbf{Training-based layout control.}
A common direction is to inject explicit spatial conditioning during training. Methods fine-tune pretrained backbones~\cite{SDXL,SD3,FLUX,chen2023pixartalpha,chen2024pixartdelta,Wuerstchen} or add adapters with grounding/alignment objectives~\cite{tokencompose,jiang2024comat,ELLA}, and layout-controlled variants train conditioning modules for boxes, masks, or keypoints~\cite{gligen,controlnet,instancediffusion,2024ranni,2023reco,creatilayout,lin2025ctrladapter,localcontrol,Zheng_2023_CVPR,gani2024llm,T2Iadapter,controlnet_plus_plus}.
These approaches enforce geometry but require substantial training and often degrade coherence and realism under hard constraints~\cite{2024realcompo}.

\noindent\textbf{RL-based compositional control.}
Reinforcement learning optimizes explicit rewards for counts and spatial relations~\cite{flowgrpo,dancegrpo,ddpo,dpok,mixgrpo,tempflowgrpo}. Such rewards improve compositional scores but induce distribution shift: models may exploit the reward at the expense of realism, even with KL regularization. In contrast, we remain inference-only and avoid reward-driven distribution shift.

\noindent\textbf{Training-free (inference-only) control.}
Attention- and latent-based methods edit text embeddings or cross-attention to steer local content~\cite{attend_excite,2023structurediff,2024contrstdiff,2024objattrbind,2023divandbind,2023linguistbind,2022composablediff,Agarwal_2023_ICCV,Gong_2024_CVPR} and perform region-wise denoising~\cite{RPG,li2024mulan}.
Layout-driven training-free techniques use LLM-derived 2D layouts (boxes/blobs) to modulate attention or emphasize regions~\cite{GPT-4o,layoutgpt,gligen,blobgen,2024realcompo,2023boxdiff,boundedattention,2023attnmodulation,2024directeddiff,Couairon_2023_ICCV,2024wacvvedaldi,2024compositediff,attnrefocus}.
These preserve base-model quality better than training-based approaches, but control is weaker and brittle for unusual layouts, high object counts, and 3D relations because guidance is limited to \emph{coarse 2D} signals without appearance priors.
Zero-Painter~\cite{zeropainter} takes user-specified masks and per-object prompts to generate objects separately and then inpaint the background, resulting in a flat 2D assembly of frozen objects that fails on overlaps. 
In contrast, we generate layouts from text and compose a depth-ordered 2.5D prior that guides full-canvas denoising, via a dual-guidance: \emph{object-prior reinforcement} anchors early denoising to preserve appearance and depth ordering, while \emph{spatially controlled denoising} strengthens spatial placement.

\noindent\textbf{Inference-time noise search \& initialization.}
A complementary thread exploits diffusion's sensitivity to initialization, either by searching over seeds and selecting the best sample~\cite{ma2025inference,ReflectDIT,CARINOX,FMTT} or by optimizing the noise/trajectory directly~\cite{inito,instancegen}. These improve success rates but are compute-intensive and brittle for out-of-distribution, highly compositional prompts; we avoid costly stochastic exploration by using a single structured prior.
Image editing similarly leverages noisy initialization or inversion~\cite{2022sdedit,2022blendeddiff,2023guidedinitialedit}. We instead \emph{generate} the initialization as \emph{composite object priors} encoding appearance and coarse 2.5D structure.

\section{The Proposed Method}
\label{sec:method}
As illustrated in Figure~\ref{fig:framework}, our \emph{ComposeAnything} framework consists of three key components for compositional text-to-image generation:
1) 2.5D Layout Planning (Section~\ref{sec:method_llm}): We employ LLMs to transform the input prompt into a structured 2.5D semantic layout, including object captions, bounding boxes and relative depths;
2) Composite Object Prior (Section~\ref{sec:method_prior}): Based on the layout, we generate a coarse composite image that serves as a strong semantic and spatial prior for guiding image synthesis;
and 3) Prior Guided Diffusion (Section~\ref{sec:method_scontrol}): We iteratively initialize noises with the object prior and apply spatial-controlled self-attention to preserve structure in early denoising steps.

\subsection{2.5D Layout Planning}
\label{sec:method_llm}

We leverage LLM planning to derive a 2.5D composite prior with spatial, appearance, and depth cues. Since recent LLMs can generate high-quality scene layouts from text~\cite{layoutgpt, RPG, hu2024scenecraft}, we use GPT-4.1~\cite{GPT-4o} to produce a structured 2.5D semantic layout from the prompt, explicitly predicting \emph{relative depth ordering} and \emph{appearance-rich object captions} that are later materialized into an RGB composite prior.

This layout consists of object captions $\{y_{o_i}\}_{i=1}^K$ (size, orientation, appearance), 2D bounding boxes $\{box_i\}_{i=1}^K$, and depth values $\{depth_{i}\}_{i=1}^K$  that encode relative depth ordering and enable 3D-aware composition, together with a background caption $y^{bg}$ and a global compositional caption $y^{global}$. The layout is obtained via a chain-of-thought reasoning process. Details are provided in the Supplementary Material. Because the generated layout is interpretable in image space, it enables efficient human-in-the-loop refinement; simple edits can substantially improve final generation, as shown in Table~\ref{tab:3dspatial_pair_a}.

\begin{figure*}[t]
    \centering
    \includegraphics[width=0.70\linewidth]{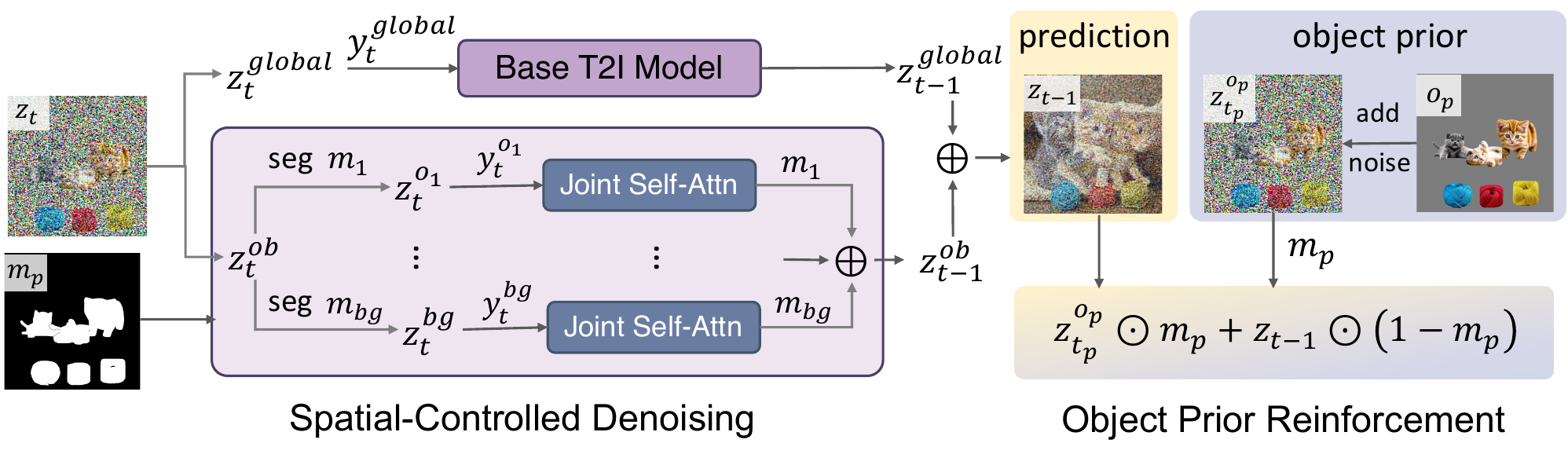}
    \caption{Overview of prior-guided diffusion. Spatial-controlled denoising is applied for each aligned text and region pair to strengthen spatial control. We further re-inject the object prior $z^{o_p}_{t_p}$ into predicted $z_{t-1}$ to reinforce the prior.}
    \label{fig:PGD}
    \vspace{-1em}
\end{figure*}

\subsection{Composite Object Prior}
\label{sec:method_prior}

\paragraph{2.5D position-aware composite image generation.}
Given the isolated object captions from LLM, we first generate individual objects using a base text-to-image diffusion model, \eg, Stable Diffusion-3 Medium (SD3-M)~\cite{SD3}. 
Next, we use a referring expression segmentation model Hyperseg~\cite{hyperseg} to extract objects $\{o_i\}_{i=1}^K$ along with their segmentation masks $\{m_i\}_{i=1}^{K}$. 
Each object and its corresponding mask are resized to fit within the designated bounding box generated from the LLM according to a scaling factor $scale_i$. 
Objects are then composited in a depth-aware order, where objects with smaller depth values are placed above those with larger depth values, thereby establishing occlusion-correct layering in the final scene. This process is formulated as follows: \( o_i', m_i' = \operatorname{Resize}(o_i, m_i, \mathit{scale}_i) \); 
Finally, all objects are composited on a $N \times N$ sized canvas, denoted as $o_p$. \( o_p, m_p = \operatorname{Compose}(\{o_i'\}, \{m_i'\}, \{box_i\}, \{depth_i\}) \).
Its corresponding composited mask is denoted as $m_{p}$.
Figure~\ref{fig:framework} shows an example of the 2.5D composite image and mask (three cats and balls of yarn). The composition of all objects forms the foreground, and the rest is considered the background. Background is freely generated via the background caption $y^{bg}$.

\paragraph{Initializing object prior for diffusion-based models.}
Our method enhances existing T2I diffusion models to generate images with complex object compositions. It is compatible with both denoising diffusion probabilistic models like SDXL \cite{SDXL} and recent flow-matching based models like SD3-M~\cite{SD3}.

The core idea of diffusion models is to learn a generative process by simulating and then reversing a gradual noising procedure.
Given an image $x_0 \sim p(x)$, the forward (diffusion) process progressively corrupts $x_0$ with Gaussian noise according to a predefined noise schedule, producing $x_t$ at timestep $t$ and yielding an approximately standard normal latent at the terminal time, i.e., $x_T \approx \mathcal{N}(0, I)$:
\begin{equation}
x_t = \alpha(t)x_0 + \sigma(t)z, \quad z \sim \mathcal{N}(0, I),
\label{eqn:diff_forward}
\end{equation}
where $t \in [0,T]$ indexes the diffusion timestep. A denoising network $\epsilon(\theta)$ is trained to predict the added noise at each step in the forward process. 
During inference, image generation starts from pure Gaussian noise $x_T$ and denoises it back to $x_0$ via the reverse process, which is an ordinary differential model (ODE) on time \(t \in [T, 0]\) guided by the noise prediction network \(\epsilon(\theta)\): \(x_{t - \Delta t} \gets x_t \;-\; \epsilon(\theta)(x_t, t)\,\Delta t.\)

Instead of starting from pure Gaussian noise at $t = T$, we initialize the process with a noisy object prior at an intermediate timestep $t_p < T$, providing a stronger starting point for generation. Specifically, we follow latent diffusion models~\cite{SD} where the denoising is applied on the latent space.
We use the above composite image $o_p$ to generate an initial noise in the latent space. 
The image $o_p$ is first encoded through a Variational Autoencoder (VAE) to get the prior latent. 
Then, we apply the forward process from Eq.~(\ref{eqn:diff_forward}) at a high noise timestep $t_p$ to obtain the latent object prior, $z^{o_p} = \text{VAE}(o_p)$, and its noised version $\hat{z}_{t_p}^{o_p} = \alpha(t_p)z^{o_p} + \sigma(t_p)z$, with $z \sim \mathcal{N}(0,I)$. Since the background in $o_p$ is empty, we avoid conditioning the generation process on the uninformative background region of the latent $\hat{z}_{t_p}^{o_p}$. 
To achieve this, we use the mask $m_p$ to reinitialize the background with pure Gaussian noise: \( z_{t_p}^{o_p} = \hat{z}_{t_p}^{o_p} \odot m_p + z_{bg} \odot (1 - m_p) \),
where $z_{bg} \sim \mathcal{N}(0, I)$ and $\odot$ indicates element multiplication. 
This ensures that only the object regions are guided by a prior, while the background remains free to be generated based on the background caption $y^{bg}$.
The reverse process still starts from $t=T$, but uses the composite object prior $z_{t_p}^{o_p}$ as initialization.

\subsection{Prior-guided Diffusion}
\label{sec:method_scontrol}

Because the 2.5D composite prior consists of layered foreground RGB cutouts without a coherent background, naively denoising from this sparse initialization often leads to two failure modes: the model either washes out foreground semantics or fails to synthesize a coherent scene context around the objects.
To address this, we introduce \emph{prior-guided diffusion}, which makes the composite prior effective by \emph{factorizing} guidance into two complementary roles (Figure~\ref{fig:PGD}).
\emph{Spatial-Controlled Denoising} stabilizes fine-grained region-caption alignment and enables both object and background synthesis conditioned on their corresponding textual descriptions.
\emph{Object Prior Reinforcement} anchors the denoising trajectory to preserve foreground appearance, semantic identity, and occlusion-based depth ordering.
Once the global structure and object relationships are established in the early diffusion steps, both guidance mechanisms are removed. The process then reverts to standard denoising, allowing the model to recover fine details and high photorealistic quality.

\paragraph{Spatial-controlled denoising.}
\label{subsec:SCD}

\begin{figure*}[!t]
  \centering
  \includegraphics[width=0.80\textwidth]{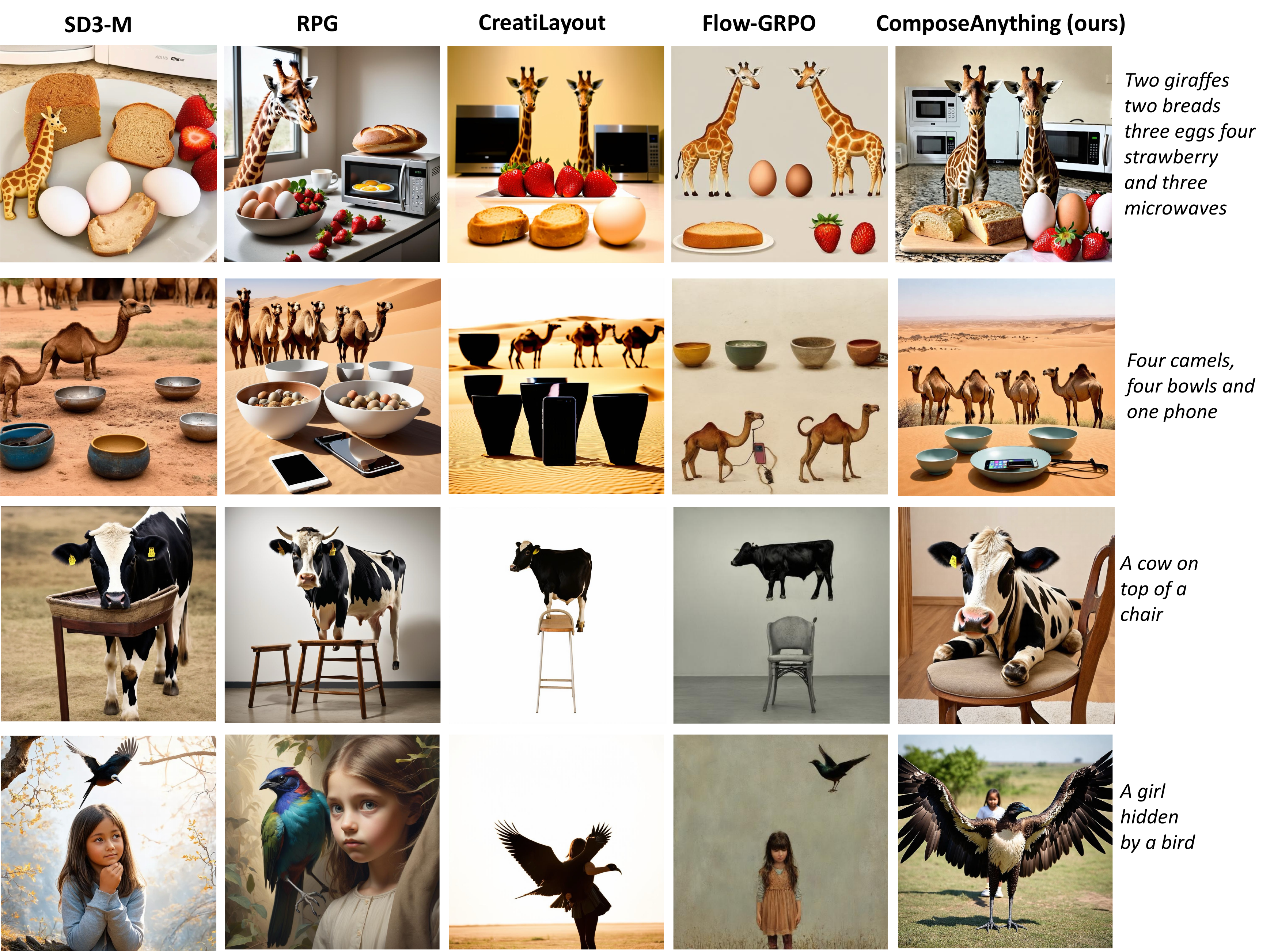}
  \caption{State-of-the-art comparison on counting, 2D-spatial, and 3D-spatial prompts against inference-only and training-based baselines.}
  \label{fig:sota_cmpr}
  \vspace{-1em}
\end{figure*}

Our method builds on Multi-Modal Diffusion Transformers, a dual-stream architecture used in Stable Diffusion-3~\cite{SD3}, which processes text and image modalities in parallel.
In addition to the global prompt embeddings $y^{global}$, we introduce a set of $K$ object prompt embeddings $\{y^{o_i}\}_{i=1}^K$ and one background prompt embedding $y^{bg}$. 
These are independently processed by the text stream, while the image stream receives only the latent image embeddings.

During the self-attention, the image latent $z_t$ is split into two latents: 
1) a global latent $z^{global}_t$, and 2) an object-background latent $z^{ob}_t$. 
Given object masks $\{m_i\}_{i=1}^K$ and background mask $m_{bg}$, we segment $z^{ob}_t$ into separate objects and background latents:
\begin{equation*}
\{z_t^{o_i}\}_{i=1}^K = \text{Seg}(z_t^{ob}, \{m_i\}_{i=1}^K), \quad
z_t^{bg} = \text{Seg}(z_t^{ob}, m_{bg})
\end{equation*}

Each object latent $z_t^{o_i}$ and its corresponding prompt embedding $y^{o_i}_t$ are concatenated and passed through a Joint Self-Attention (JSA) module:
\begin{equation*}
\begin{aligned}
q_t^i &= \bigl[(W_{qy} \cdot y_t^{o_i});\,(W_{qz} \cdot z_t^{o_i})\bigr], \\
k_t^i &= \bigl[(W_{ky} \cdot y_t^{o_i});\,(W_{kz} \cdot z_t^{o_i})\bigr], \\
v_t^i &= \bigl[(W_{vy} \cdot y_t^{o_i});\,(W_{vz} \cdot z_t^{o_i})\bigr].
\end{aligned}
\end{equation*}
\begin{equation*}
\bigl[y_t^{o_i}, z_t^{o_i}\bigr]
\;\leftarrow\;
\text{Softmax}\!\left(\frac{q_t^i (k_t^i)^\top}{\sqrt{d}}\right) \cdot v_t^i
\end{equation*}
where $W_{{\cdot y}}$ project prompt embeddings and $W_{{\cdot z}}$ project image latents. For simplicity, we reuse the same notation for the input and output of the transformer layer.
This spatial-controlled self-attention is applied at each transformer layer, enabling precise control over object placement and appearance.
The same mechanism is applied to the background: $[y_{t}^{bg}, z_{t}^{bg}] \leftarrow \text{JSA}(y_t^{bg}, z_t^{bg})$. 
The original global attention is applied on the global prompt and the global latent embeddings $[y_{t}^{global}, z_{t}^{global}] \leftarrow \text{JSA}(y_t^{global}, z_t^{global})$. 

After the last transformer layer, the object and background latents are denoised from $t \rightarrow{} t-1$.
The updated object latents $\{z_{t-1}^{o_i}\}_{i=1}^K$ and background latent $z_{t-1}^{bg}$ are then composed back into $z_{t-1}^{ob}$ using the segmentation masks.

Finally, we merge the global latent and object-background latent with a weighted sum:
\( z_{t-1} = z_{t-1}^{global} * \text{ratio}_{global} + z_{t-1}^{ob} * (1 - \text{ratio}_{global}). \) It balances global coherence from the global latent and fine-grained spatial control from the object-background latent, which allows global consistency.
We apply the spatial control from time $t=T$ until the latent reaches time $t_{sc}$.
A smaller $t_{sc}$ corresponds to a stronger spatial control.

\paragraph{Object prior reinforcement.}
To avoid overly corrupting the foreground prior, we initialize the foreground from a \emph{partially noised} version of the composite prior at timestep $t_p$ (i.e., by applying the forward diffusion process up to $t_p$). 
In contrast, the background is initialized from \emph{fully random} terminal noise at $T$, i.e., $x_T \sim \mathcal{N}(0,I)$.
However, during denoising from $t=T$, this mismatch in noise levels leads to inaccurate noise predictions for the foreground region, which potentially distort its semantics and structure.
To address this, we propose the foreground prior reinforcement algorithm.
During the denoising steps from $T$ to $t_p$, we repeatedly restore the original object prior in the foreground regions to protect them from degradation.
Specifically, we overwrite the foreground region in the current latent $z_{t-1}$ with the initial object prior, while retaining the denoised background: \( z_{t-1} \leftarrow z_{t_p}^{o_p} \odot m_p + z_{t-1} \odot (1 - m_p). \)

This iterative replacement ensures that the semantic correctness and spatial structure of the object prior are preserved throughout the early diffusion steps. Meanwhile, the background is progressively refined in the presence of a fixed foreground, allowing coherent integration between the two.

Once the latent reaches time $t_p$, both foreground and background are aligned in terms of noise level and the global structure becomes stable.
From this point onward, denoising proceeds without any additional intervention, allowing for natural refinement and generative flexibility. 
Notably, decreasing $t_{p}$ strengthens the object prior while reducing generative flexibility.

\begin{table*}[!t]
\centering
\sbox{\tabOne}{%
   \footnotesize\setlength{\tabcolsep}{8pt}\renewcommand{\arraystretch}{1.0}%
  \begin{tabular}{@{}lcccccc@{\hspace{14pt}}cc@{}}
    \toprule
    \multirow{2}{*}{Method}
      & \multicolumn{6}{c}{T2I-CompBench} & \multicolumn{2}{c}{NSR-1k} \\
    \cmidrule(lr){2-7}\cmidrule(l){8-9}
    & 2D-Spatial & Count & 3D-Spatial & Complex & Color & Texture & Spatial & Count \\
    \midrule
    SD-v1~\cite{SD}                 & 12.46 & 44.61 & --    & 30.80 & 37.65 & 41.56 & 16.89 & 31.45 \\
    Attend-Excite v2~\cite{attend_excite}
                                    & 14.55 & 47.67 & --    & 34.01 & 64.00 & 59.63 & 26.86 & 39.41 \\
    SDXL~\cite{SDXL}                & 21.33 & 49.88 & 47.12 & 32.37 & 58.79 & 52.99 & 31.57 & 30.62 \\
    PixArt-$\alpha$~\cite{chen2023pixartalpha}    & 20.82 & 50.58 & 38.62 & 33.57\rlap{$^{*}$} & 68.86 & 70.44 & -- & --\\    
    RealCompo~\cite{2024realcompo}  & 31.73 & 65.92 & --    & 31.27\rlap{$^{*}$}    & 77.41 & \textbf{74.27} & --    & --    \\
    SD3-M~\cite{SD3}                & 31.32 & 60.22 & 49.43 & 37.71 & 81.32 & 73.34 & 44.43 & 44.61 \\
    FLUX~\cite{FLUX}                & 26.13 & 60.58 & 59.51 & 37.03 & 74.07 & 69.22 & 39.29 & 55.97 \\
    RPG$^{*}$~\cite{RPG}            & 40.26 & 56.39 & 50.93 & 36.53 & 76.80 & 61.95 & 52.24 & 39.81 \\
    RAGD$^{*}$~\cite{RAGD}          & 42.56 & 57.38 & 62.42 & 36.68 & 80.39 & 70.85 & --    & --    \\
    Inference-scale~\cite{ma2025inference}
                                    & 31.51 & 67.89 & --    & 38.10 & \textbf{83.03} & 73.64 & --    & --    \\
    \midrule
    ComposeAnything                 & \textbf{48.24} & \textbf{68.21} & \textbf{77.16} & \textbf{38.66} & 81.11 & 73.74 & \textbf{63.80} & \textbf{59.36} \\
    \bottomrule
  \end{tabular}%
}
\captionsetup{width=\wd\tabOne}
\caption{Quantitative comparison of inference-based methods on T2I-CompBench and NSR-1k. $^{*}$ denotes our reproduced result with their released code.}
\label{tab:sota_results}
\usebox{\tabOne}
\end{table*}

\begin{table*}[!t]
\centering
\sbox{\tabGen}{%
  \footnotesize\setlength{\tabcolsep}{3pt}\renewcommand{\arraystretch}{1.0}%
  \begin{tabular}{@{}lcccccc>{\columncolor[HTML]{FBFAEF}}c>{\columncolor[HTML]{FBF4EF}}c@{}}
    \toprule
    Method & \makecell{Single\\obj.} & \makecell{Two\\obj.} & Count & Color & Pos. & Attr. & Overall & \makecell{Avg.\\time} \\
    \midrule
    SD3-M~\cite{SD3}        & 0.99 & 0.84 & 0.56 & 0.84 & 0.32 & 0.52 & 0.68 & \textbf{4.0} \\
    RPG$^{*}$~\cite{RPG}    & 0.99 & 0.82 & 0.43 & 0.76 & 0.63 & 0.64 & 0.71 & 15.49 \\
    CARINOX~\cite{CARINOX}  & 1.00 & 0.86 & 0.66 & 0.90 & 0.16 & 0.48 & 0.68 & 70.0 \\
    Reflect-DIT~\cite{ReflectDIT} & 0.98 & 0.96 & 0.80 & 0.88 & 0.66 & 0.60 & 0.81 & 118.0 \\
    InstanceGen~\cite{instancegen} & 0.97 & 0.87 & 0.68 & 0.85 & 0.64 & 0.73 & 0.79 & 300.0 \\
    ComposeAnything         & 1.00 & 0.93 & 0.70 & 0.85 & 0.83 & 0.72 & \textbf{0.84} & 16.37 \\
    \bottomrule
  \end{tabular}%
}
\sbox{\tabQual}{%
  \footnotesize\setlength{\tabcolsep}{2.2pt}\renewcommand{\arraystretch}{1.0}%
  \begin{tabular}{@{}lcccc@{\hspace{7pt}}cc@{}}
    \toprule
    & \multicolumn{4}{c}{Compositionality} & \multicolumn{2}{c}{Image quality} \\
    \cmidrule(lr){2-5}\cmidrule(l){6-7}
    Method & 2D-Sp. & Count & 3D-Sp. & Cplx. & Aesth.$\uparrow$ & HPSv3$\uparrow$ \\
    \midrule
    SD3-M base~\cite{SD3}              & 31.32 & 60.22 & 49.43 & 37.71 & 5.40 & 6.89 \\
    CreatiLayout~\cite{creatilayout} & 47.36 & 62.15 & 68.85 & 34.60 & 5.26 & 4.22 \\
    Flow-GRPO~\cite{flowgrpo}        & \textbf{54.47} & 67.52 & 48.49 & 38.42 & 5.42 & 5.99 \\
    \midrule
    \textbf{Ours} & 48.24 & \textbf{68.21} & \textbf{77.16} & \textbf{38.66} & \textbf{5.47} & \textbf{7.06} \\
    \bottomrule
  \end{tabular}%
}
\noindent\begin{minipage}[t]{\wd\tabGen}
  \centering\captionsetup{width=\linewidth}
  \captionof{table}{Quantitative comparison on GenEval. $^{*}$ denotes reproduced results; average inference time over GenEval is in seconds.}
  \label{tab:geneval_tab}\usebox{\tabGen}
\end{minipage}%
\hfill%
\begin{minipage}[t]{\wd\tabQual}
  \centering\captionsetup{width=\linewidth}
  \captionof{table}{Compositionality and image quality for training based methods on T2I-CompBench. }
  \label{tab:quality_compbench}\usebox{\tabQual}
\end{minipage}
\end{table*}

\section{Experiments}
\label{sec:experiments}
\subsection{Experimental Setup}

\noindent\textbf{Evaluation benchmarks.}
We evaluate our method on three commonly used benchmarks T2I-CompBench~\cite{huang2023t2icompbench}, NSR-1K~\cite{layoutgpt} and GenEval~\cite{geneval}. 
They contain prompts rich in spatial, 3D, numeric, generally complex and surreal compositions.
 We focus on six categories from T2I-CompBench: 2D Spatial, Numeracy (Count), 3D Spatial, Complex, Color, and Texture, each containing 300 prompts.
For NSR-1K, we report results on their two categories, Spatial (283 prompts) and Count (672 prompts). 
GenEval evaluates T2I models on complex compositional prompts across 6 different categories including spatial, counting, and attributes with a total of 553 prompts.

\noindent\textbf{Evaluation metrics.}
For 2D-spatial and counting in T2I-CompBench and NSR-1K, we follow the standard detector-based protocols for object identification, counting, and spatial relations; for color and texture, we use the standard BLIP-VQA protocol.
For 3D-spatial, the original T2I-CompBench metric relies on outdated depth/detection models, so we introduce a GPT-4.1~\cite{GPT-4o} metric aligned with recent MLLM evaluation practice~\cite{zhang2023gpt}: it checks required objects and their 3D relations, then averages normalized 0--100 scores.
We use the T2I-CompBench 3-in-1 metric for Complex and the standard GenEval detection/segmentation and CLIP-based attribute protocols; additional metric details are in the Supplementary Material.

\noindent\textbf{Implementation details.}
We use GPT-4.1~\cite{GPT-4o} for 2.5D layout planning without manual edits and SD3-Medium (SD3-M)~\cite{SD3} as the default base model.
Generation uses 28 denoising steps and two complementary timestep hyperparameters: $t_p$ controls how strongly the foreground prior anchors appearance and occlusion-based depth, while $t_{sc}$ controls how long spatial constraints stabilize global layout and context.
Decoupling them preserves prior structure while allowing the model to recover global coherence; based on a held-out T2I-CompBench subset, we use $t_p=0.91$ and $t_{sc}=0.96$ unless otherwise specified.

\subsection{Comparison to state of the art}

\begin{figure*}[!t]
  \centering
  \includegraphics[width=\textwidth]{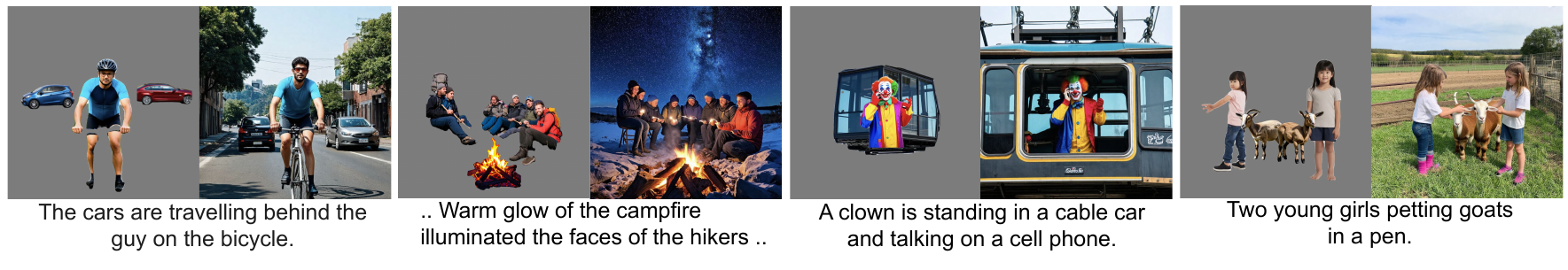}
  \vspace{-2em}
  \caption{\textbf{Robustness to priors} (\textit{left}: prior, \textit{right}: final). \textbf{(1)} Missing bicycle, recovered with correct shadows; \textbf{(2)} no firelight in prior, faces correctly lit; \textbf{(3)} awkward clown placement, refined coherently; \textbf{(4)} incoherent poses, natural object interaction \& correct camera angle.}
  \label{fig:rebuttal_prior_guided_diffusion}
\end{figure*}

\begin{table*}[t]
\centering
\footnotesize
\setlength{\tabcolsep}{3pt}
\renewcommand{\arraystretch}{1.05}
\begin{subtable}[t]{0.335\textwidth}\centering
\begin{tabular}{@{}cccccc@{}}
\toprule
\rule{0pt}{2.6ex}OPR & SCD & 2D & Count & 3D & Complex \\
\midrule
$\times$ & $\times$ & 31.32 & 60.22 & 49.43 & 37.71 \\
$\checkmark$ & $\times$ & 45.33 & 66.08 & \textbf{77.38} & 38.16 \\
$\times$ & $\checkmark$ & 43.56 & 64.42 & 75.48 & 37.41 \\
$\checkmark$ & $\checkmark$ & \textbf{48.24} & \textbf{68.21} & 77.16 & \textbf{38.66} \\
\bottomrule
\end{tabular}
\caption{Impact of OPR and SCD.}
\label{tab:ablation_compbench_a}
\end{subtable}
\hfill
\begin{subtable}[t]{0.30\textwidth}\centering
\begin{tabular}{@{}lcccc@{}}
\toprule
\rule{0pt}{2.6ex} & 2D & Count & 3D & Complex \\
\midrule
OP & 45.79 & 68.06 & 86.71 & 35.04 \\
FI & 48.24 & 68.21 & 77.16 & 38.66 \\
\bottomrule
\end{tabular}
\caption{Object prior (OP) vs.\ final image (FI).}
\label{tab:ablation_compbench_b}
\end{subtable}
\hfill
\begin{subtable}[t]{0.335\textwidth}\centering
\begin{tabular}{@{}lcccc@{}}
\toprule
\rule{0pt}{2.6ex}Method & VQA$\uparrow$ & TIFA$\uparrow$ & Aes.$\uparrow$ & HPSv3$\uparrow$ \\
\midrule
SD3-base & 0.930 & 0.863 & 5.265 & \textbf{11.69} \\
Flow-GRPO & 0.930 & 0.869 & 5.209 & \phantom{0}9.55 \\
CreatiLayout & 0.917 & 0.865 & 5.261 & \phantom{0}6.31 \\
\textbf{Ours} & \textbf{0.939} & \textbf{0.878} & \textbf{5.265} & 11.05 \\
\bottomrule
\end{tabular}
\caption{MS-COCO compositions and quality.}
\label{tab:quality_coco}
\end{subtable}

\vspace{0.9em}

\begin{subtable}[t]{0.335\textwidth}\centering
\begin{tabular}{@{}l@{\hspace{8pt}}c@{}}
\toprule
\rule{0pt}{2.6ex}Method & 3D-Spatial \\
\midrule
SD3-M (no prior) & 49.43 \\
RPG 2D regions & 50.93 \\
CreatiLayout 2D boxes & 68.85 \\
\midrule
Reverse layered 2.5D & 54.23 \\
Random layered 2.5D & 66.41 \\
Ours (2.5D layered) & 77.16 \\
\;\;+ manual editing & \textbf{94.27} \\
\bottomrule
\end{tabular}
\caption{2D vs.\ 2.5D priors; manual editing.}
\label{tab:3dspatial_pair_a}
\end{subtable}
\hfill
\begin{subtable}[t]{0.30\textwidth}\centering
\begin{tabular}{@{}lcccc@{}}
\toprule
\rule{0pt}{2.6ex}Method & 2D & Count & 3D & Complex \\
\midrule
SDXL & 21.33 & 49.88 & 47.12 & 32.37 \\
+ \emph{ours} & \textbf{44.64} & \textbf{57.22} & \textbf{71.03} & \textbf{36.20} \\
\midrule
SD3-M & 31.32 & 60.22 & 49.43 & 37.71 \\
+ \emph{ours} & \textbf{48.24} & \textbf{68.21} & \textbf{77.16} & \textbf{38.66} \\
\midrule
FLUX & 26.13 & 60.58 & 59.51 & 37.03 \\
+ \emph{ours} & \textbf{44.21} & \textbf{67.83} & \textbf{76.14} & \textbf{37.36} \\
\bottomrule
\end{tabular}
\caption{Generalization across base models.}
\label{tab:ablation_base_models}
\end{subtable}
\hfill
\begin{subtable}[t]{0.335\textwidth}\centering
\begin{tabular}{@{}lccccc@{}}
\toprule
\rule{0pt}{2.6ex}& \multicolumn{4}{c}{Single GPU} & Multi \\
\cmidrule(lr){2-5}\cmidrule(l){6-6}
\#Obj. & 3 & 5 & 7 & 10 & 10 \\
\midrule
LLM plan.   & 3.50 & 4.15 & 5.33 & 6.94 & 6.94 \\
Object gen. & 12.0 & 19.0 & 27.0 & 38.0 & \textbf{4.0} \\
Segmentation       & 0.6  & 0.7  & 1.0  & 1.2  & \textbf{0.5} \\
Image gen.  & 5.1  & 5.4  & 5.9  & 6.1  & 6.1 \\
\midrule
Total       & 21.2 & 29.3 & 39.2 & 52.2 & \textbf{17.5} \\
\bottomrule
\end{tabular}
\caption{Inference time (s) by stage.}
\label{tab:runtime}
\end{subtable}
\caption{Ablation studies on T2I-CompBench (MS-COCO for quality) and runtime analysis.}
\label{tab:ablations}
\end{table*}

\noindent\textbf{Comparison with inference-based methods.}
We compare against pretrained T2I models, layout-guided training-free methods, inference-scaling/noise-search methods, and latent-optimization baselines listed in Tables~\ref{tab:sota_results} and~\ref{tab:geneval_tab}; coverage differs across benchmarks due to incomplete reporting in prior work. 
In Tab.~\ref{tab:sota_results}, our method achieves the strongest object-level structural scores (2D-Spatial, 3D-Spatial, Count, and Complex), while remaining competitive on color and texture.
RPG and RAGD are reproduced with the same LLM planner GPT-4.1 and using their default SDXL and FLUX backbones respectively. Table \ref{tab:ablation_base_models} shows ComposeAnything results with SDXL and FLUX as backbones.
Figure~\ref{fig:sota_cmpr} further shows that baselines struggle with challenging spatial relations (e.g., a cow on top of a chair) and exact counts (e.g., giraffes and microwaves), whereas our guidance improves prompt following while preserving global coherence. We provide human evaluations against RPG in the Supplementary Material, where raters consistently prefer our results.

Table~\ref{tab:geneval_tab} reports GenEval performance and average inference time: our method is comparable in runtime to RPG~\cite{RPG} while substantially improving performance, and avoids the heavy iterative optimization or best-of-$N$ sampling costs of InstanceGen~\cite{instancegen}, CARINOX~\cite{CARINOX}, and Reflect-DIT~\cite{ReflectDIT} by using a single guided generation pass.

\noindent\textbf{Comparison with training-based methods.}
Training-based methods such as Flow-GRPO~\cite{flowgrpo} and CreatiLayout~\cite{creatilayout} outperform inference-time methods on some compositional categories, but often shift the image distribution and degrade visual quality, producing artifacts such as floating objects, faded backgrounds, and broken physics.
Table~\ref{tab:quality_compbench} and Figure~\ref{fig:sota_cmpr} show this trade-off: our inference-only prior guidance achieves the best Aesthetic~\cite{aestheticsscore} and HPSv3~\cite{hpsv3} scores while preserving coherent images under difficult compositional prompts. Additional human evaluations are in the Supplementary Material.

\begin{table}[h]\centering\scriptsize
\setlength{\tabcolsep}{2.2pt}
\begin{tabular}{@{}lccc@{\hspace{6pt}}cccccc@{}}
\toprule
& \multicolumn{3}{c}{Layout planning} & \multicolumn{6}{c}{Final image generation} \\
\cmidrule(lr){2-4}\cmidrule(l){5-10}
Planner & 2D & 3D & Num. & 2D. & Num. & 3D. & Cplx. & Color & Tex. \\
\midrule
GPT-4.1     & 99.1 & 100.0 & 98.6 & 48.24 & 68.21 & 77.16 & 38.66 & 81.11 & 73.74 \\
Qwen3.6-27B & 98.9 & 100.0 & 98.2 & 47.40 & 70.33 & 77.84 & 38.23 & 79.62 & 71.68 \\
\bottomrule
\end{tabular}
\vspace{-0.5em}
\caption{Layout-plan accuracy with image generation performance.}
\vspace{-1em}
\label{tab:planner}
\end{table}

\subsection{Ablations}

\noindent\textbf{Impact of Prior guided Diffusion.}
Table~\ref{tab:ablation_compbench_a} isolates the contribution of object-prior reinforcement (OPR) and spatial-controlled denoising (SCD). Compared with SD3-M~\cite{SD3}, OPR improves 2D, Count, 3D, and Complex by 14.01, 5.86, 27.95, and 0.45 points, respectively; SCD gives similar structural gains of 12.24, 4.20, and 26.05 points on 2D, Count, and 3D.
Combining OPR and SCD gives the best overall trade-off. 
OPR grounds depth-aware foreground structure by preserving appearance and occlusion ordering, while SCD locks the overall placement and suppresses object leakage or extra instances. Together they yield larger gains than either component alone.

\begin{figure}[h]
  \centering
  \includegraphics[width=\columnwidth]{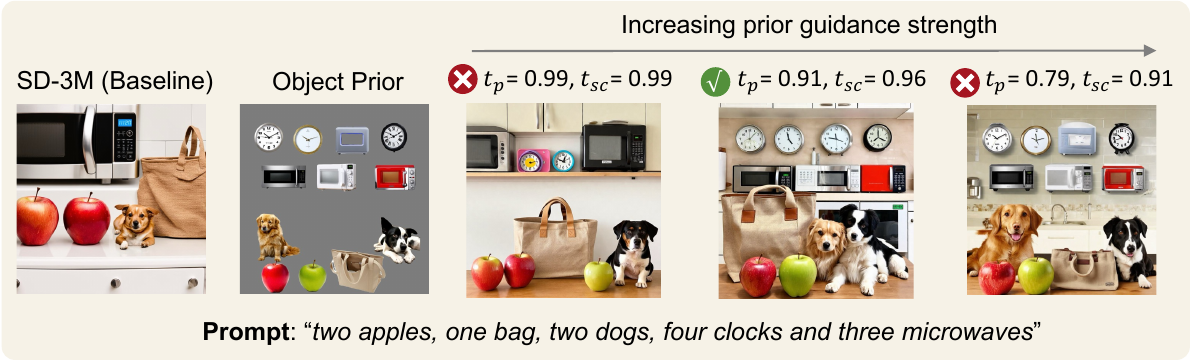}
  \caption{Effect of Object prior reinforcement and Spatial controlled Denoising strength. The baseline (left) without prior fails to satisfy the high-object-count prompt. Integrating our object prior with varying guidance strengths enables a controlled trade-off between strict adherence to the prior and the seamless generative blending by the base model.}
  \label{fig:opr_scd_strength}
  \vspace{-1em}
\end{figure}

\noindent\textbf{Strength of prior guided diffusion.}
Prior-guided diffusion uses $t_p$ for OPR and $t_{sc}$ for SCD to balance prior adherence with the model's native denoising.
Lower strength values (higher $t_p$ and $t_{sc}$) grant the diffusion model greater flexibility to modify the layout, whereas higher strength values (lower $t_p$ and $t_{sc}$) cause the prior to exert stronger control.
Figure~\ref{fig:opr_scd_strength} shows that weak guidance can drift from the intended layout, while overly strong guidance transfers the prior too rigidly and limits correction of prior errors. Intermediate strength best preserves structure while retaining enough flexibility for coherent synthesis.

\noindent\textbf{Robustness of Object Priors.}
The prior serves as structural guidance rather than a pixel-level template. This is because OPR operates only in the high-noise range ($t_p{=}0.91$), where just coarse structure is set; SCD keeps a global stream denoising the entire image with $y^{\text{global}}$, filling in what the prior misses; and the remaining 91\% of steps are unconstrained, allowing coherent lighting, perspective, poses, and natural interactions. Minor prior imperfections and segmentation misses therefore do not propagate to the final image (Fig.~\ref{fig:rebuttal_prior_guided_diffusion}).

\begin{figure}[t]
  \centering
  \includegraphics[width=0.47\textwidth]{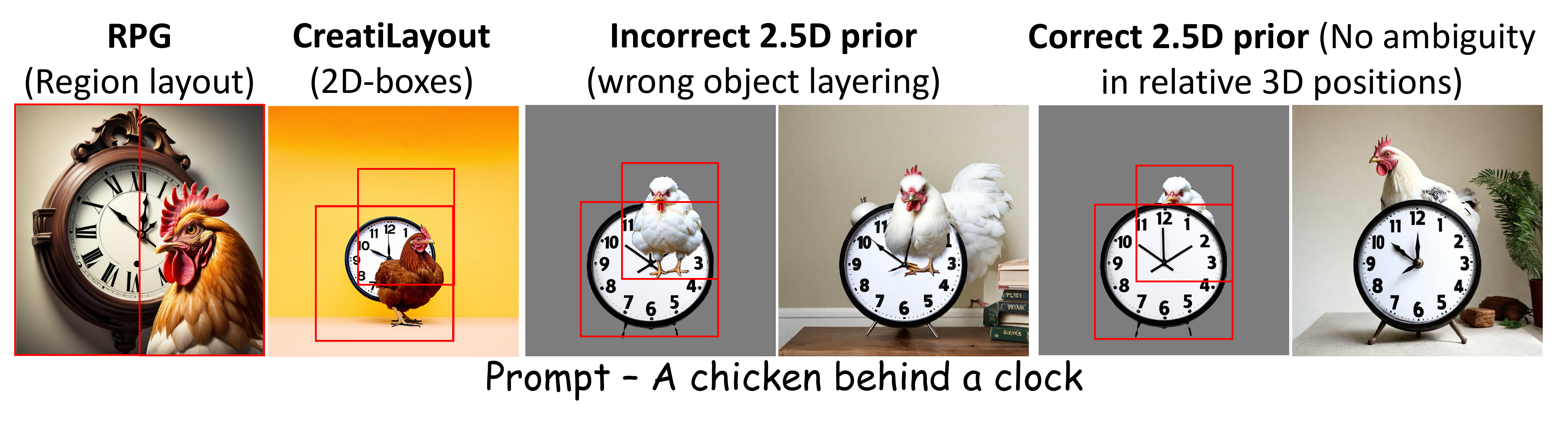}
  \vspace{-1em}
  \caption{Effect of the 2.5D prior on 3D-spatial prompt generation.}
  \label{fig:2.5d_prior}
  \vspace{-1em}
\end{figure}

\noindent\textbf{Impact of 2.5D prior.} 
Prior works~\cite{RPG,creatilayout} rely on 2D bounding-box layouts that constrain only planar positions, leaving depth relationships to the diffusion model's internal prior. As a result, inference-only methods such as RPG~\cite{RPG} and training-based approaches such as CreatiLayout~\cite{creatilayout} struggle with OOD 3D spatial relationships (Fig.~\ref{fig:2.5d_prior}).

In contrast, our 2.5D RGB priors explicitly encode occlusion and front/back depth ordering, which is particularly beneficial for 3D-Spatial prompts. Table~\ref{tab:3dspatial_pair_a} shows that our automatically generated 2.5D priors outperform 2D-layout baselines; randomizing the depth order degrades performance, and reversing it performs even worse. These results highlight the importance of accurate depth encoding for complex 3D spatial relationships.

\noindent\textbf{Manual prior editing.} 
Because prior quality correlates with final quality, manual refinement offers interpretable human-in-the-loop control. 
We adjusted 48 challenging 3D-Spatial examples from T2I-CompBench that failed in image generation with a score of 0. Edits are done on the corresponding implausible priors, such as objects being placed too close to each other or differing drastically in size.
The spatial edits include resizing and adjusting the objects such that the 3D spatial relations become visually explicit. This yields a 17.11-point gain in Table~\ref{tab:3dspatial_pair_a}. Remaining failures involve extreme scale disparities in OOD prompts, where the base model enlarges or moves tiny objects forward, conflicting with occlusion.

\noindent\textbf{Generalization to natural captions:}
To show that \emph{ComposeAnything} preserves the base SD3 distribution on human-written captions, we evaluate 300 random MS-COCO captions. Tab.~\ref{tab:quality_coco}: our method slightly improves prompt following (+0.009 VQA~\cite{vqascore}, +0.015 TIFA~\cite{tifa}) and matches SD3 on Aesthetics~\cite{aestheticsscore} and HPSv3~\cite{hpsv3} (Fig.~\ref{fig:rebuttal_prior_guided_diffusion}). In contrast, training-based Flow-GRPO~\cite{flowgrpo} and CreatiLayout~\cite{creatilayout} reduce HPSv3 due to task-specific distribution shift.

\noindent\textbf{Generalization over base models.}
We apply our method to architecturally diverse models spanning diffusion and flow-matching: U-Net based SDXL~\cite{SDXL} and Transformer-based SD3-M~\cite{SD3} and FLUX~\cite{FLUX}, using the same object prior and layout. Tab.~\ref{tab:ablation_base_models} shows significant gains over every base model, indicating the method is architecture-agnostic.

\noindent\textbf{Layout planning with open-weight LLM:} We build a ground-truth layout evaluation set for the three categories with deterministic layout rules: 2D-Spatial, 3D-Spatial, and Numeracy of T2I-Compbench. Tab.~\ref{tab:planner} reports layout accuracy and the final-image scores for GPT-4.1 and an open-weight Qwen3.6-27B \cite{qwen3}. Both plan layouts near-perfectly, and their final-image scores stay within 2 points on every category. The method is  not bottlenecked by a proprietary LLM.

\noindent\textbf{Runtime breakdown.} Tab.~\ref{tab:runtime} reports stage-wise time as the object count grows. Object generation dominates on a single GPU (12-38s for 3-10 objects), but objects are independent, so distributing them across GPUs makes this stage constant at 4s. 10 objects take 52.2s on one GPU and 17.5s in parallel.

\noindent\textbf{Error Analysis.}
Upstream errors are rare: layout planning is near-saturated (Tab.~\ref{tab:planner}). For prior construction, mostly each object is generated alone, where SD3 exceeds 99\% on GenEval single-object (Tab.~\ref{tab:geneval_tab}), and segmentation is applied to the same single-object images--an easy case for SOTA models. Tab.~\ref{tab:ablation_compbench_b} further shows that object-prior (OP) correctness is strongly correlated with final-image (FI) correctness. To quantify failures, human judges labeled 90 prior-image pairs (30 each for 2D-Spatial, 3D, and Numeracy from T2I-CompBench); a prior is incorrect if implausible (extreme size discrepancies, unrealistic 3D relations) even when the layout is satisfied. 26\% of priors are incorrect, yet 63\% of these still yield correct final images, confirming the robustness above. The remaining failures concentrate in extreme compositions (e.g., 10+ objects in Numeracy). The prior is a coarse 2D composite that the diffusion model must turn into a coherent scene, and it uses its generative freedom to do so. When the prior becomes highly unrealistic, the adjustments needed for a coherent image come at the cost of composition accuracy. Detailed analysis in Supp.\ Sec.~\ref{sec:supp_prior_image_eval}, Figs.~\ref{fig:barplot}--\ref{fig:prior_image_human_eval_2d}.

\section{Conclusion}
\label{sec:conclusion}

In this work, we introduce {\em ComposeAnything}, an inference-time framework for compositional text-to-image generation that leverages \emph{2.5D composite object priors}. 
By initializing diffusion from an interpretable object-level composite instead of geometric layouts, and guiding denoising to stay consistent with it, our method improves multi-object compositions, including counts, spatial relations, complex and novel arrangements, without any additional training.
The two mechanisms, Object Prior Reinforcement and Spatial-Controlled Denoising, allow for faithful transfer of the prior by preserving object appearance and depth ordering while keeping each region aligned with its description.
Across T2I-CompBench, NSR-1K, and GenEval, {\em ComposeAnything} achieves a state-of-the-art quality compositionality trade-off, maintaining high visual fidelity while improving prompt adherence under complex and surreal compositions.

{\small
\paragraph{Acknowledgements.}
This work was granted access to the HPC resources of IDRIS under the allocation AD011014688R1 made by GENCI. It was funded in part by the French government under management of Agence Nationale de la Recherche as part of the ``France 2030'' program, reference ANR-23-IACL-0008 (PR[AI]RIE-PSAI project), and Paris \^Ile-de-France R\'egion in the frame of the DIM AI4IDF. Cordelia Schmid would like to acknowledge the support by the K\"orber European Science Prize.
}

{
    \small
    \bibliographystyle{ieeenat_fullname}
    \bibliography{main}
}

\clearpage
\appendix
\twocolumn[{%
  \null\vskip .375in
  \begin{center}
    {\Large \bf ComposeAnything: Composite Object Priors\\for Text-to-Image Generation\\{\large Supplementary Material} \par}
    \vspace*{24pt}
  \end{center}
}]

This supplementary material provides additional analyses, qualitative results, and implementation details for \emph{ComposeAnything}, which initializes text-to-image diffusion from an editable 2.5D composite object prior and transfers it through Object Prior Reinforcement (OPR) and Spatial-Controlled Denoising (SCD).
We analyze the key hyperparameters in Section~\ref{sec:supp_hyperparams}, showing how the strengths of prior reinforcement and spatial control balance prior adherence against generative flexibility.
Section~\ref{sec:supp_prior_image_eval} provides a human study of prior-image correctness on 2D-Spatial, 3D-Spatial, and Numeracy prompts, showing that the diffusion model can often correct imperfect priors during synthesis, together with qualitative examples of object priors and final images from different categories of T2I-CompBench.
Section~\ref{sec:supp_human_evals_screenshot} provides SOTA human evaluations against the inference-based method RPG and the training-based methods CreatiLayout and Flow-GRPO, while Sections~\ref{sec:supp_llm_prompt}-\ref{sec:supp_3d_llm_metric} give the full LLM prompts for 2.5D layout planning and for evaluating 3D-Spatial relations.

\section{Impact of Key Hyperparameters}
\label{sec:supp_hyperparams}

We decouple and analyze two key hyperparameters, $t_p$ and $t_{sc}$.

$t_p$ is the prior noising time and controls object prior reinforcement (OPR), where lower values denote stronger priors. $t_{sc}$ is the time at which spatial control is removed and controls spatial-controlled denoising (SCD), where larger values result in stronger control.

\begin{figure*}[p]
  \centering
  \includegraphics[height=0.82\textheight]{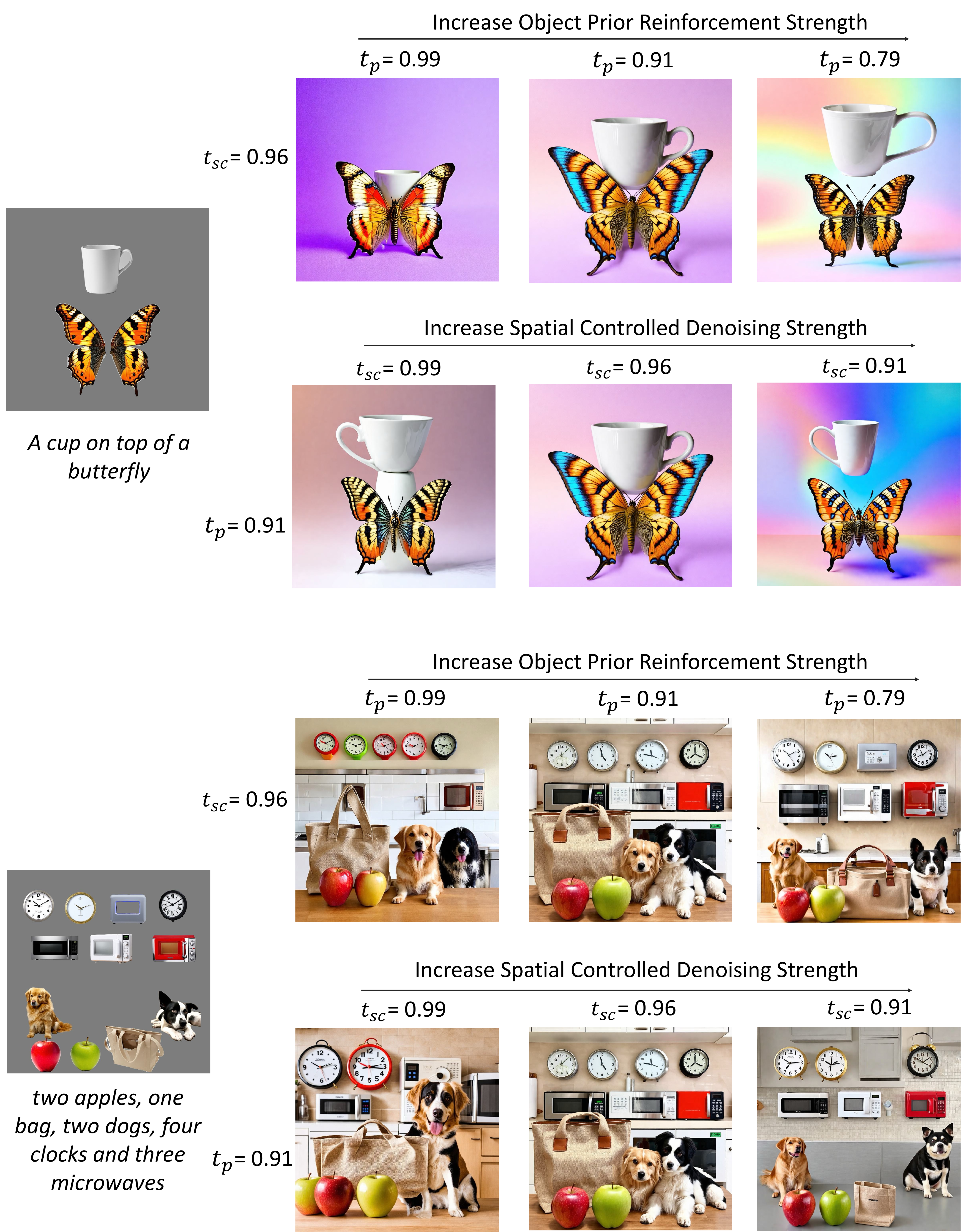}
  \caption{Effect of Object Prior Reinforcement and Spatial-Controlled Denoising. Increasing either strength enhances appearance fidelity and spatial precision, but reduces generative flexibility.}
  \label{fig:prior_scd_strength}
\end{figure*}

Figure~\ref{fig:prior_scd_strength} presents object priors and corresponding generated images for two text prompts under varying $t_p$ and $t_{sc}$ values.
In the first row of each example, $t_{sc}$ is fixed at the default value of 0.96 to isolate OPR under different $t_p$ values.
Weak OPR fails to preserve prior semantics and appearance; for example, the butterfly appears in front of the cup rather than on top of it, and the number of clocks and microwaves is incorrect.
As OPR strength increases, object semantics and appearance, including color, shape, and count, are more strongly retained.
However, excessive reinforcement reduces generative flexibility, yielding over-constrained and less natural outputs.

In the second row, $t_p$ is fixed at the default value of 0.91 to examine SCD under different $t_{sc}$ values.
Weak SCD causes limited spatial control, with objects leaking into the background (extra cup), objects merging (two dogs merged), and incorrect object counts.
As SCD strength increases, object positions and sizes from the prior are more faithfully preserved in the final image.
Overly strong control can produce rigid placements, incoherent scenes, and floating objects, similar to training-based box-conditioned methods~\cite{creatilayout}.

We set $t_p=0.91$ and $t_{sc}=0.96$ to balance prompt adherence, generative flexibility, and overall scene coherence.
Both hyperparameters are beneficial and complementary for reliably producing correct spatial relations, accurate object counts, and high-quality images.

\section{Detailed Evaluation of Object Priors and Final Images}
\label{sec:supp_prior_image_eval}To assess object priors and final images, we categorize results into four combinations:
i) correct prior-correct image,
ii) correct prior-incorrect image,
iii) incorrect prior-correct image, and
iv) incorrect prior-incorrect image.
This separation helps identify whether failures originate from prior construction or from the final image synthesis stage.
It also reveals when a correct prior is not sufficient, because synthesis must still balance realism, visual quality, and count fidelity.

\begin{figure*}[t!]
  \centering
  \includegraphics[width=0.86\textwidth]{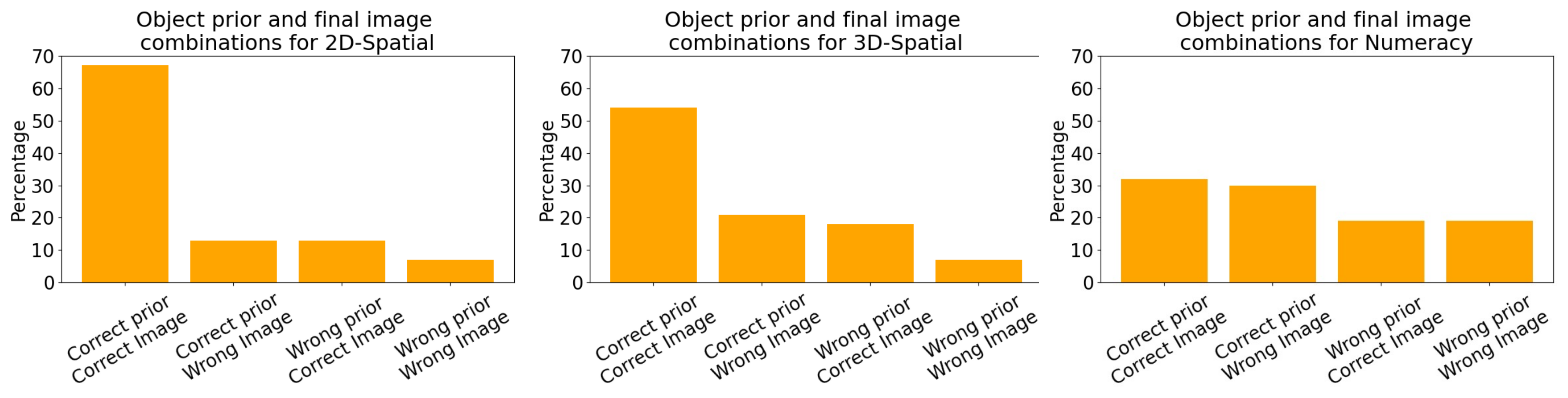}
  \caption{Human evaluation results on the correctness of object-prior and final-image pairs across three categories of T2I-CompBench.}
  \label{fig:barplot}
  \vspace{0.35em}
  \includegraphics[height=0.69\textheight]{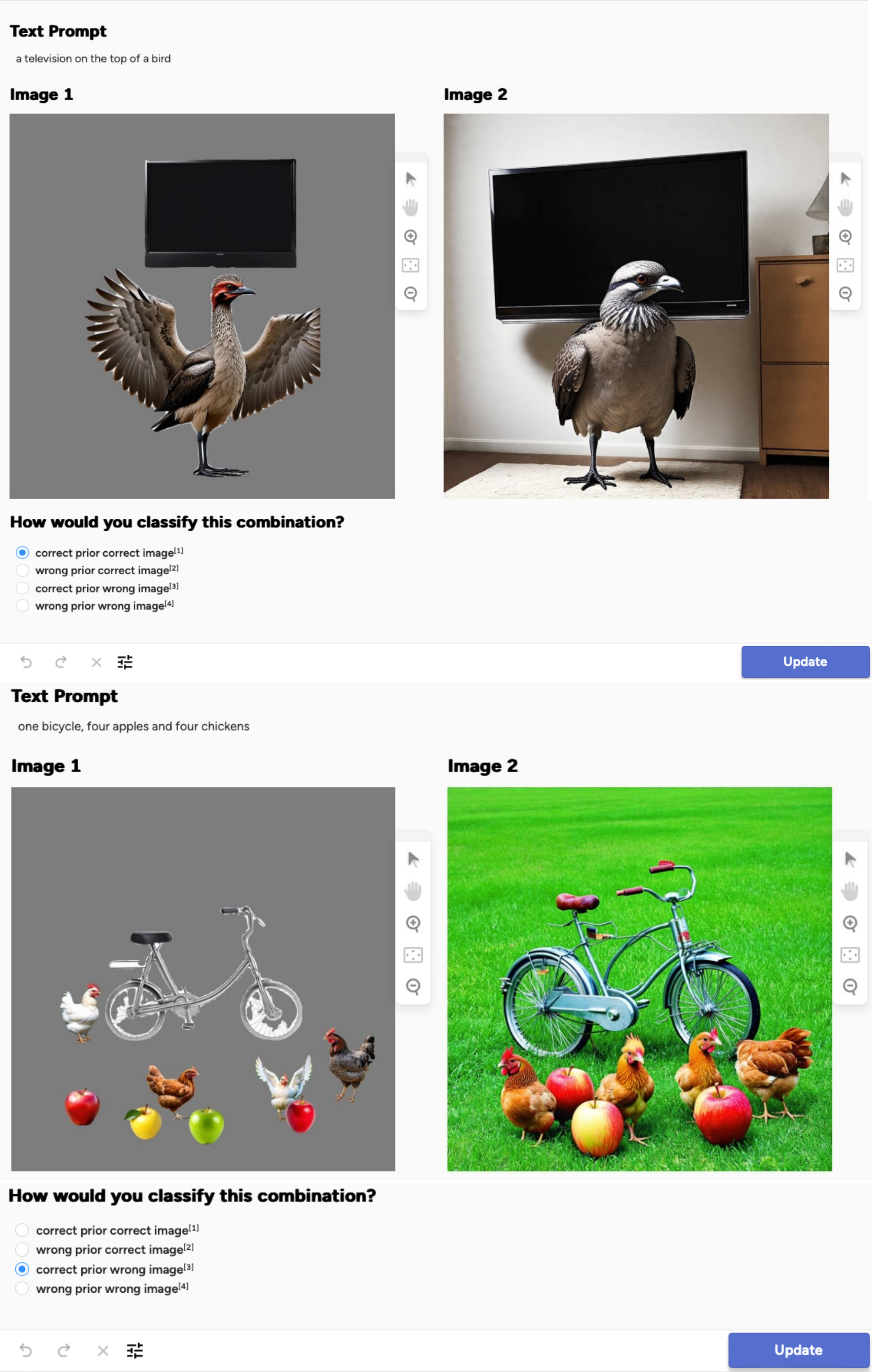}
  \caption{Labeling interface for evaluating the object prior and the final image.}
  \label{fig:prior_image_human_eval_2d}
\end{figure*}

We conduct a human evaluation using 30 prior-image pairs per category across the 2D-Spatial, 3D-Spatial, and Numeracy categories in T2I-CompBench.
Annotators perform a 4-way classification judging the correctness of both the prior and final image; the interface is shown in Figure~\ref{fig:prior_image_human_eval_2d}.

Figure~\ref{fig:barplot} shows that most 2D-Spatial and 3D-Spatial samples have both correct priors and correct images.
Numeracy has more correct-prior but incorrect-image cases, mainly because larger object counts make relative sizes affect realism and quality.
Although our model's generative flexibility helps correct such artifacts during image synthesis, it can do so at the expense of count accuracy, leading to mismatches in object quantities, as illustrated in Figure~\ref{fig:prior_image_human_eval_2d} (bottom).
Moreover, as object counts increase, priors tend to become less coherent overall, resulting in more incorrect-prior and incorrect-image cases.

\begin{figure*}[p]
  \centering
  \includegraphics[width=\textwidth]{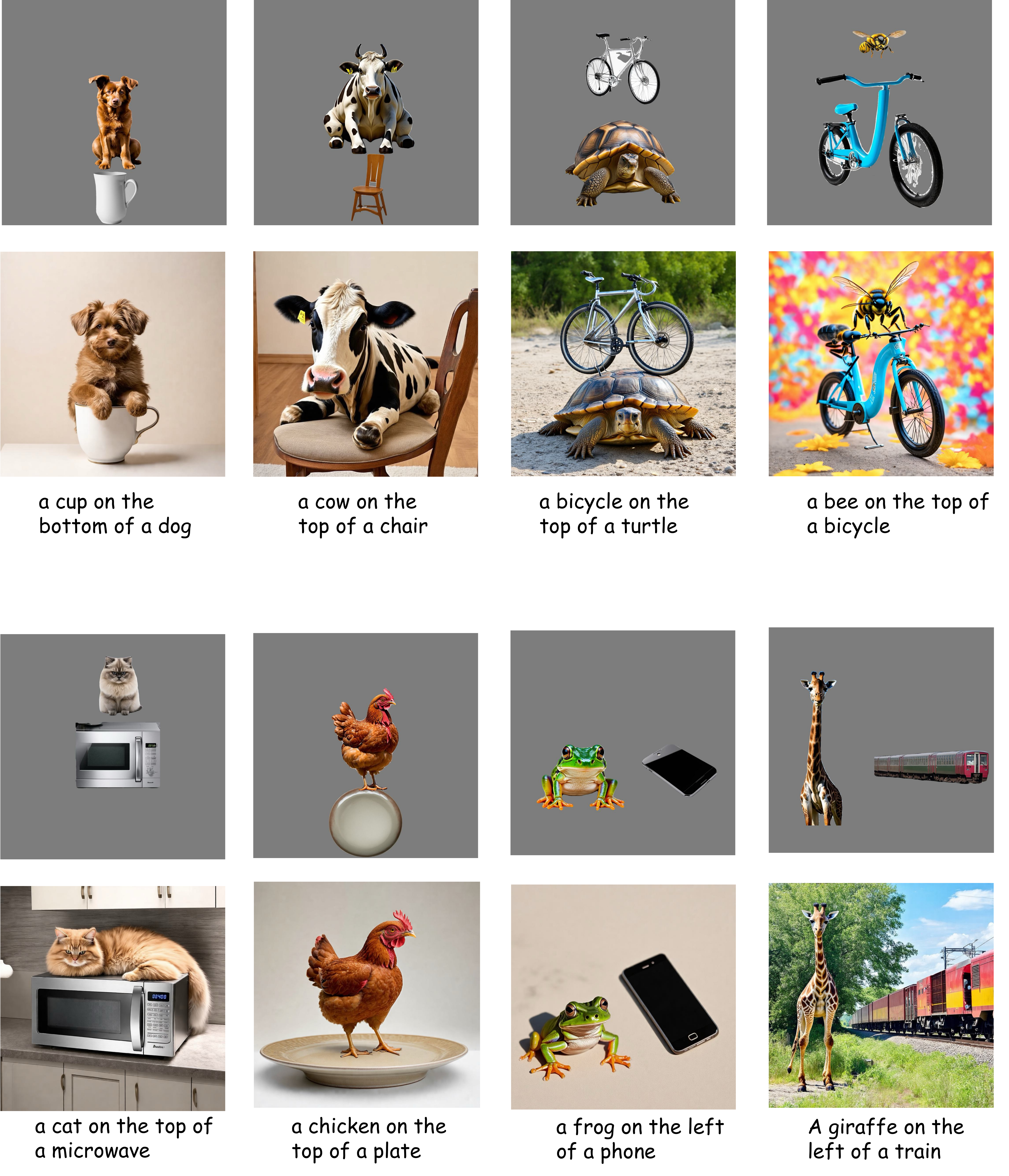}
  \caption{Object prior and corresponding generation for 2D-Spatial compositions from T2I-CompBench.}
  \label{fig:2d-spatial}
\end{figure*}

\begin{figure*}[p]
  \centering
  \includegraphics[width=\textwidth]{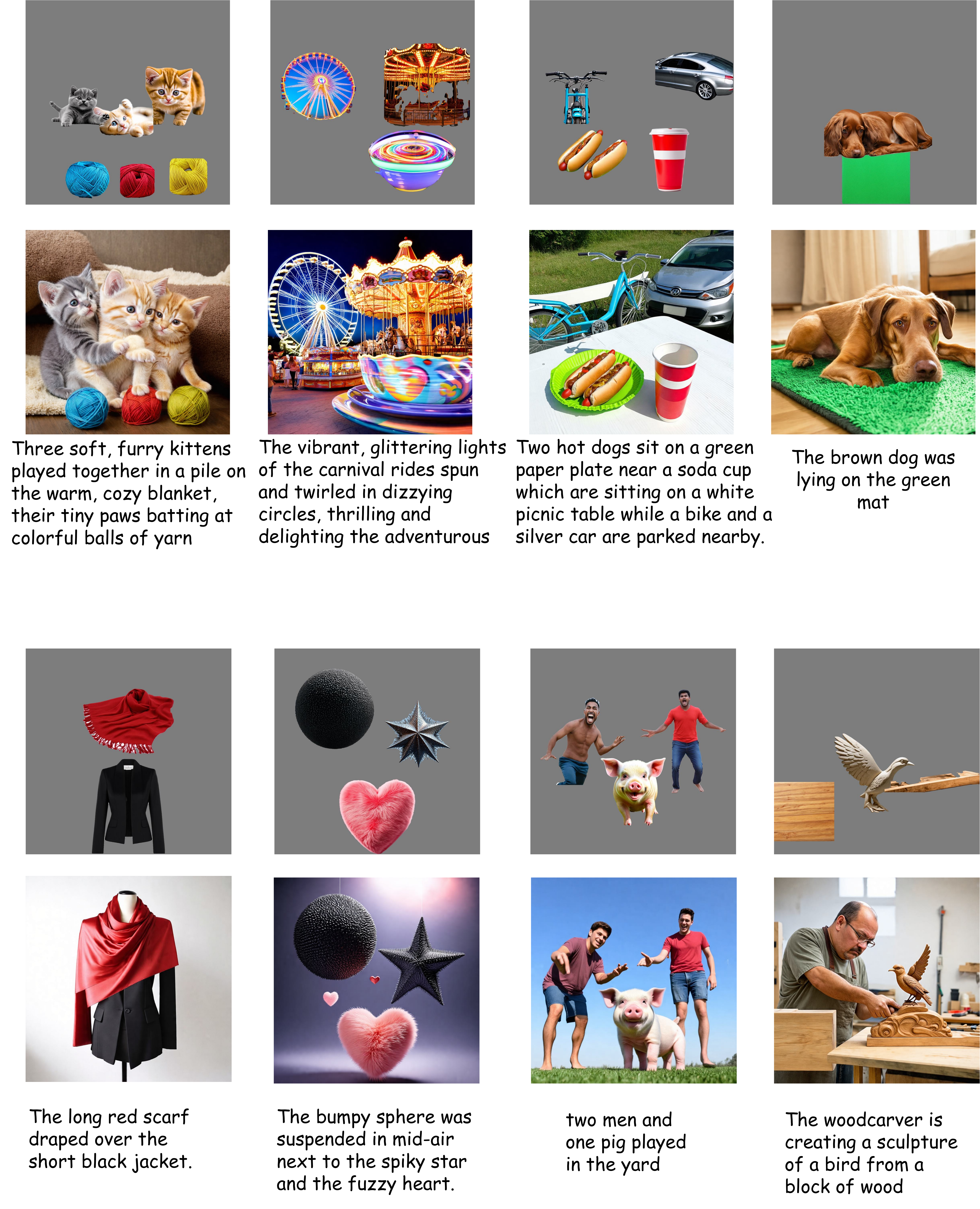}
  \caption{Object prior and corresponding generation for Complex compositions from T2I-CompBench.}
  \label{fig:complex-1}
\end{figure*}

\begin{figure*}[p]
  \centering
  \includegraphics[width=\textwidth]{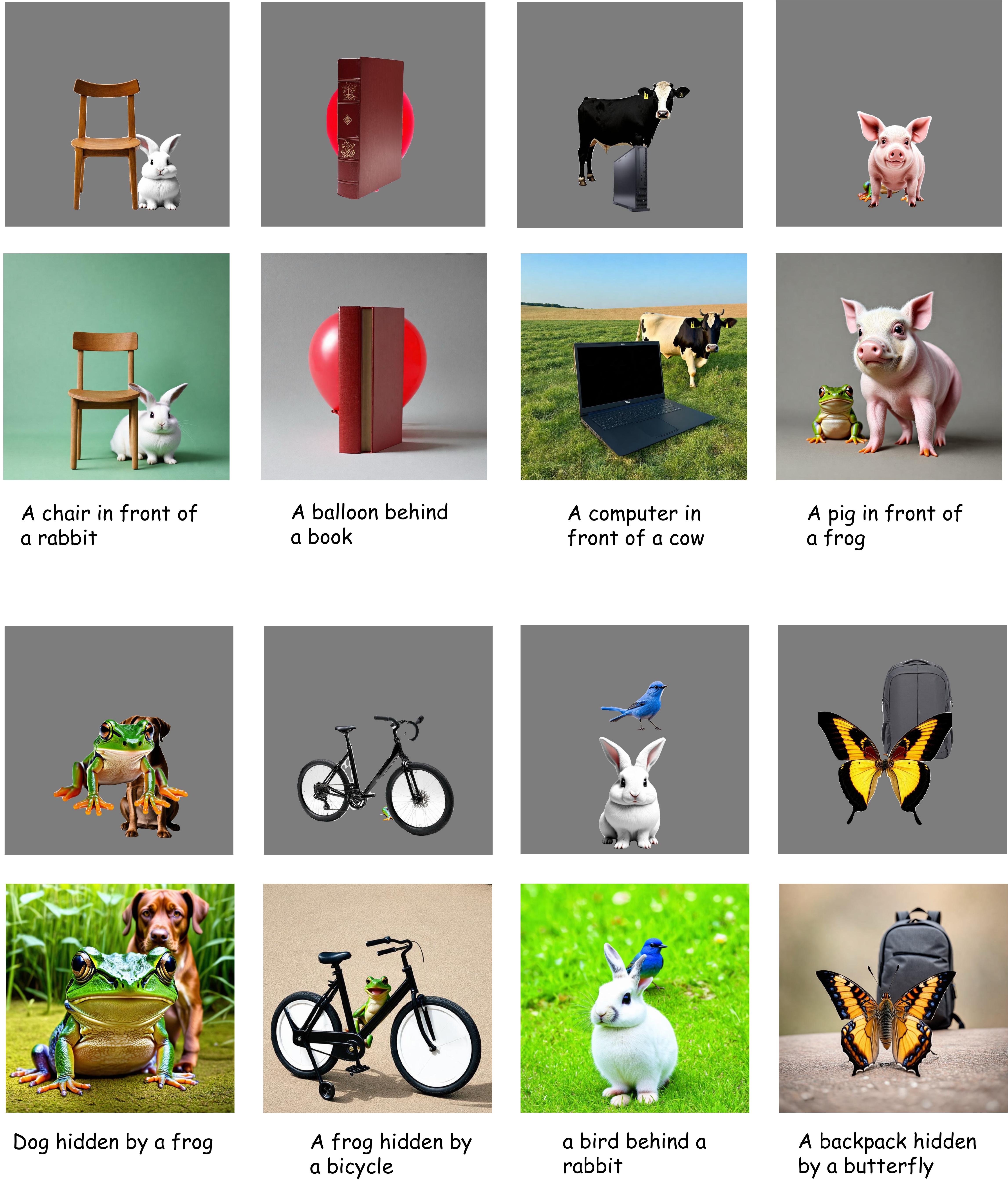}
  \caption{Object prior and corresponding generation for 3D-Spatial compositions from T2I-CompBench.}
  \label{fig:3d-spatial}
\end{figure*}

\begin{figure*}[p]
  \centering
  \includegraphics[width=\textwidth]{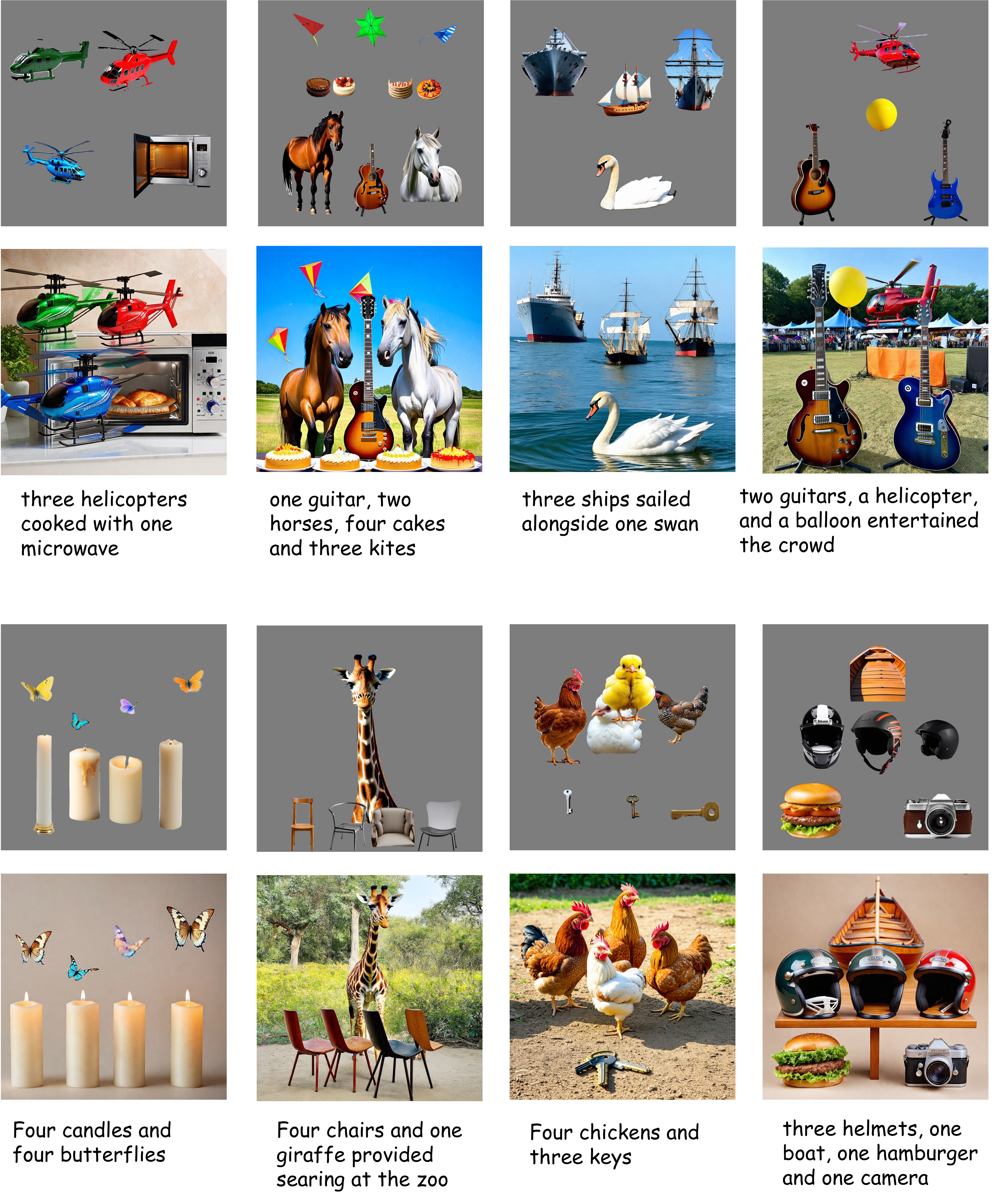}
  \caption{Object prior and corresponding generation for Numeracy compositions from T2I-CompBench.}
  \label{fig:numeracy1}
\end{figure*}

Figures~\ref{fig:2d-spatial}-\ref{fig:numeracy1} show additional examples of object priors and corresponding final images across 2D-Spatial, 3D-Spatial, Numeracy, and Complex prompt categories.

\clearpage

\clearpage

\clearpage

\clearpage

\clearpage

\clearpage

\clearpage
\twocolumn[{%
  \centering
  \includegraphics[width=0.95\textwidth]{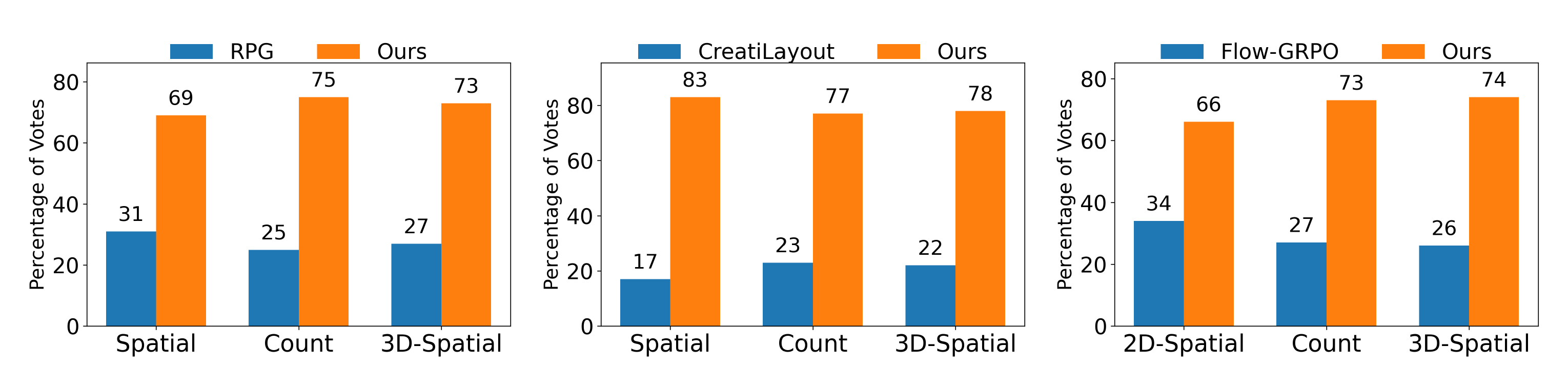}
  \captionof{figure}{Human evaluations against the inference-based method RPG~\cite{RPG} and training-based methods CreatiLayout~\cite{creatilayout} and Flow-GRPO~\cite{flowgrpo}.}
  \label{fig:humanevalsota}
  \vspace{0.75em}
}]

\section{SOTA Human Evaluations}
\label{sec:supp_human_evals_screenshot}

\begin{figure*}[p]
  \centering
  \includegraphics[height=0.82\textheight]{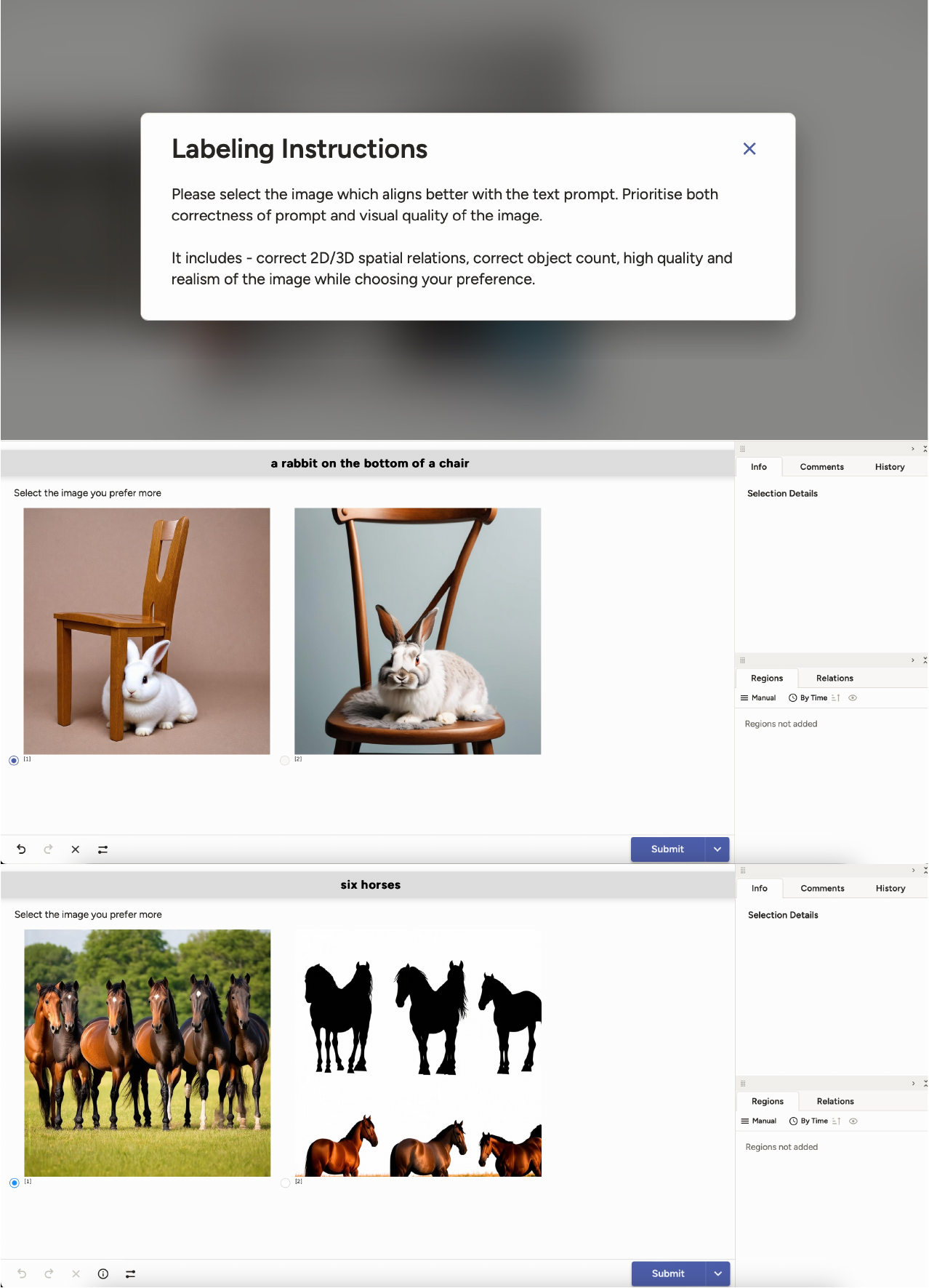}
  \caption{Labeling interface for human evaluations.}
  \label{fig:labeling_interface}
\end{figure*}

We conduct human evaluations against SOTA methods on three categories in T2I-CompBench: 2D-Spatial, 3D-Spatial, and Count.
For each category, we randomly sample 30 prompts and perform pairwise comparisons.
Five raters participated in the evaluation, with 30\% of images overlapping across raters to measure inter-annotator agreement.
The average agreement score was around 80\% for all categories.
Figure~\ref{fig:labeling_interface} shows the human evaluation instructions and interface.
The instructions focus on prioritizing correctness with respect to the prompt, particularly 2D and 3D spatial relations and object count, while also selecting images with higher quality.
As shown in the ``six horses'' example, both images contain six horses, but the first image has better visual quality.

We compare with the state-of-the-art inference-based method RPG~\cite{RPG}, and two training-based methods, CreatiLayout~\cite{creatilayout} and Flow-GRPO~\cite{flowgrpo}.
As shown in Figure~\ref{fig:humanevalsota}, human raters consistently prefer our method over RPG, CreatiLayout, and Flow-GRPO across all metrics.
This agrees with the qualitative comparisons in Figure~\ref{fig:sota_cmpr}, which highlight that inference-only baselines often fail to follow the input prompt, especially for difficult compositions.
In contrast, training-based methods can generate correct compositions, but they shift the natural image distribution to fit surreal compositions, reducing image quality.

\section{LLM Planning Instructions}
\label{sec:supp_llm_prompt}
\begin{figure*}[p]
  \centering
  \includegraphics[height=0.82\textheight]{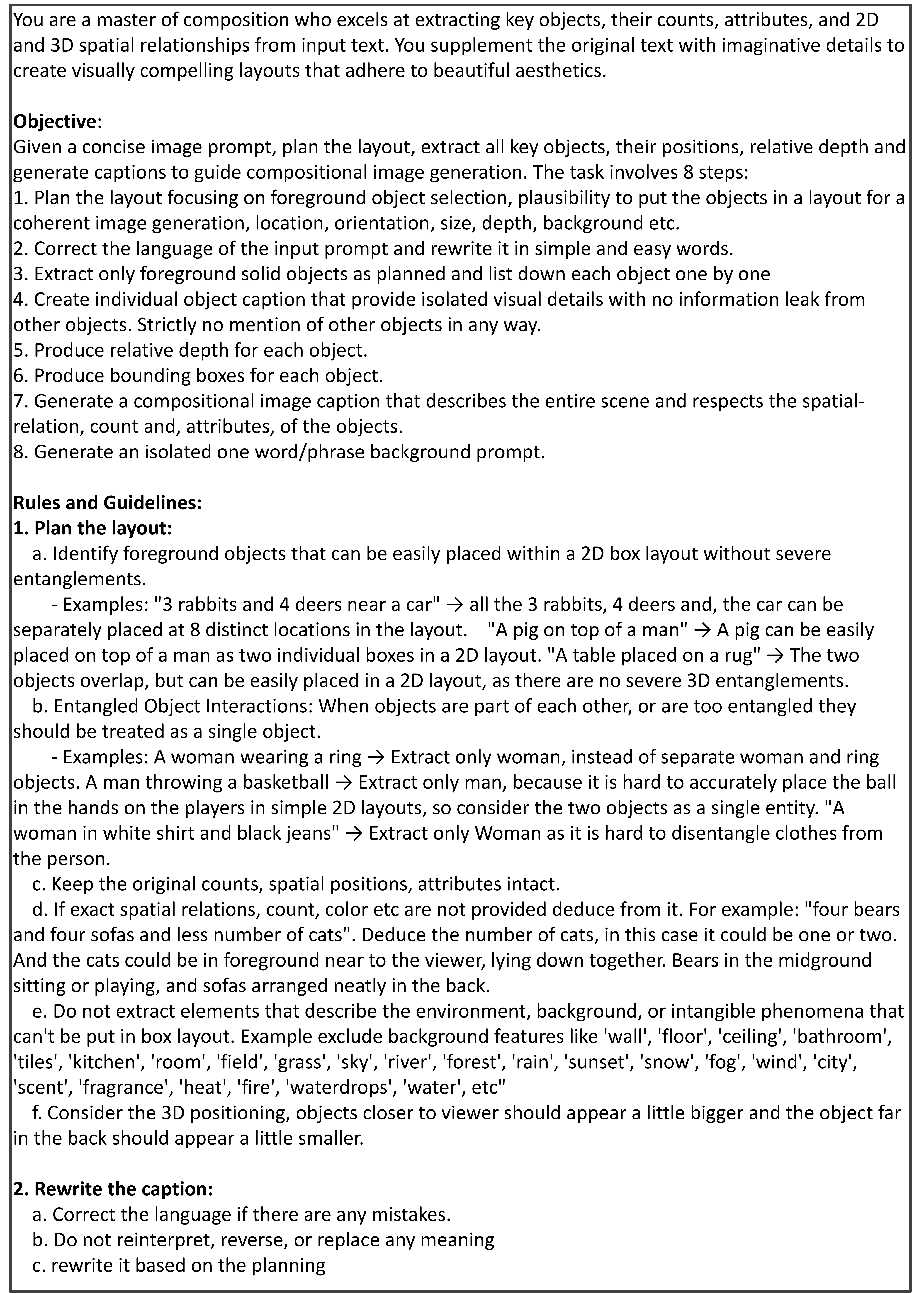}
  \caption{Instructions for LLM planning (continued in the next figure).}
  \label{fig:llm_instruct_1}
\end{figure*}

\begin{figure*}[p]
  \centering
  \includegraphics[height=0.82\textheight]{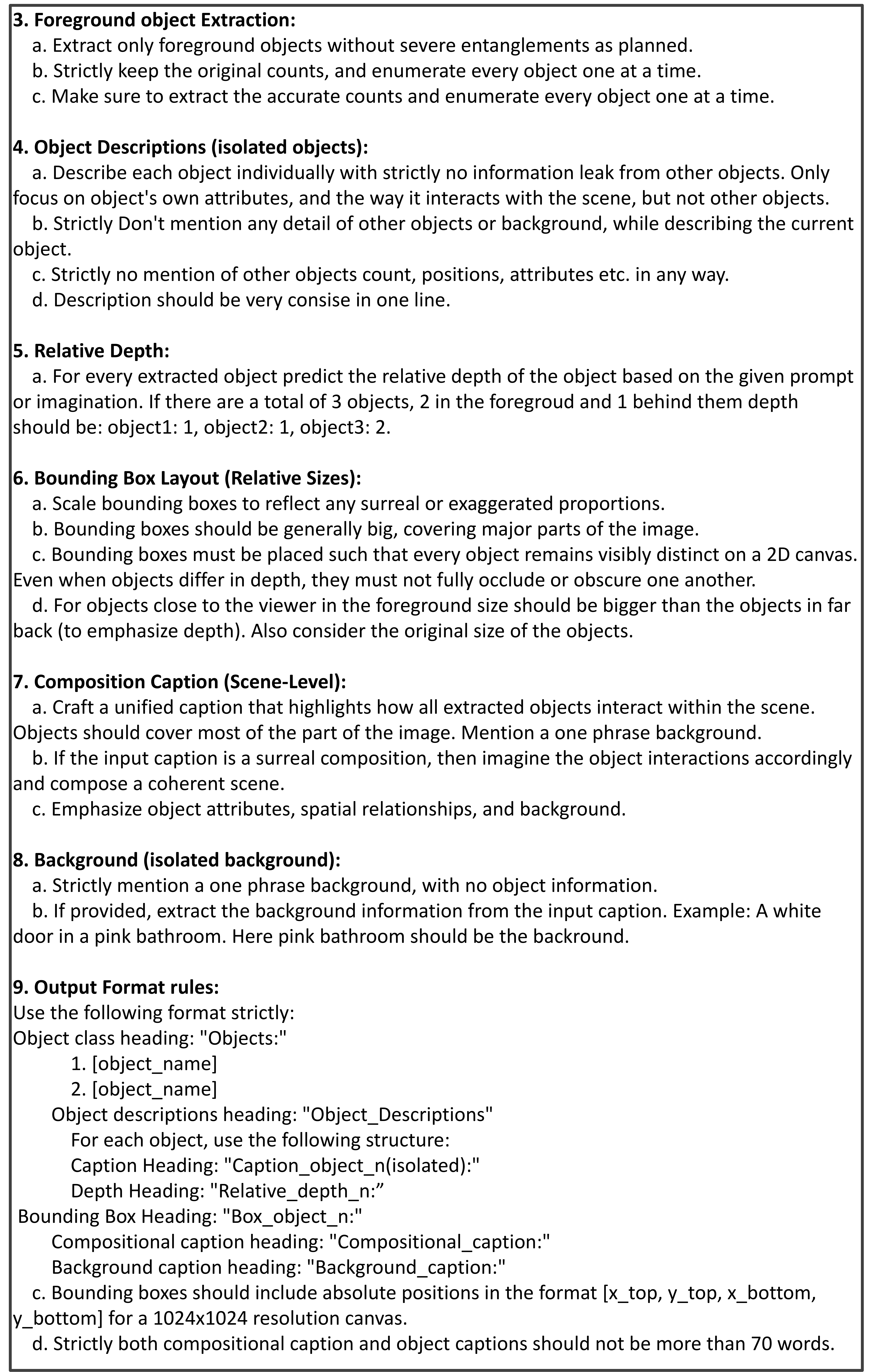}
  \caption{Instructions for LLM planning (continued in the next figure).}
  \label{fig:llm_instruct_2}
\end{figure*}

\begin{figure*}[p]
  \centering
  \includegraphics[height=0.82\textheight]{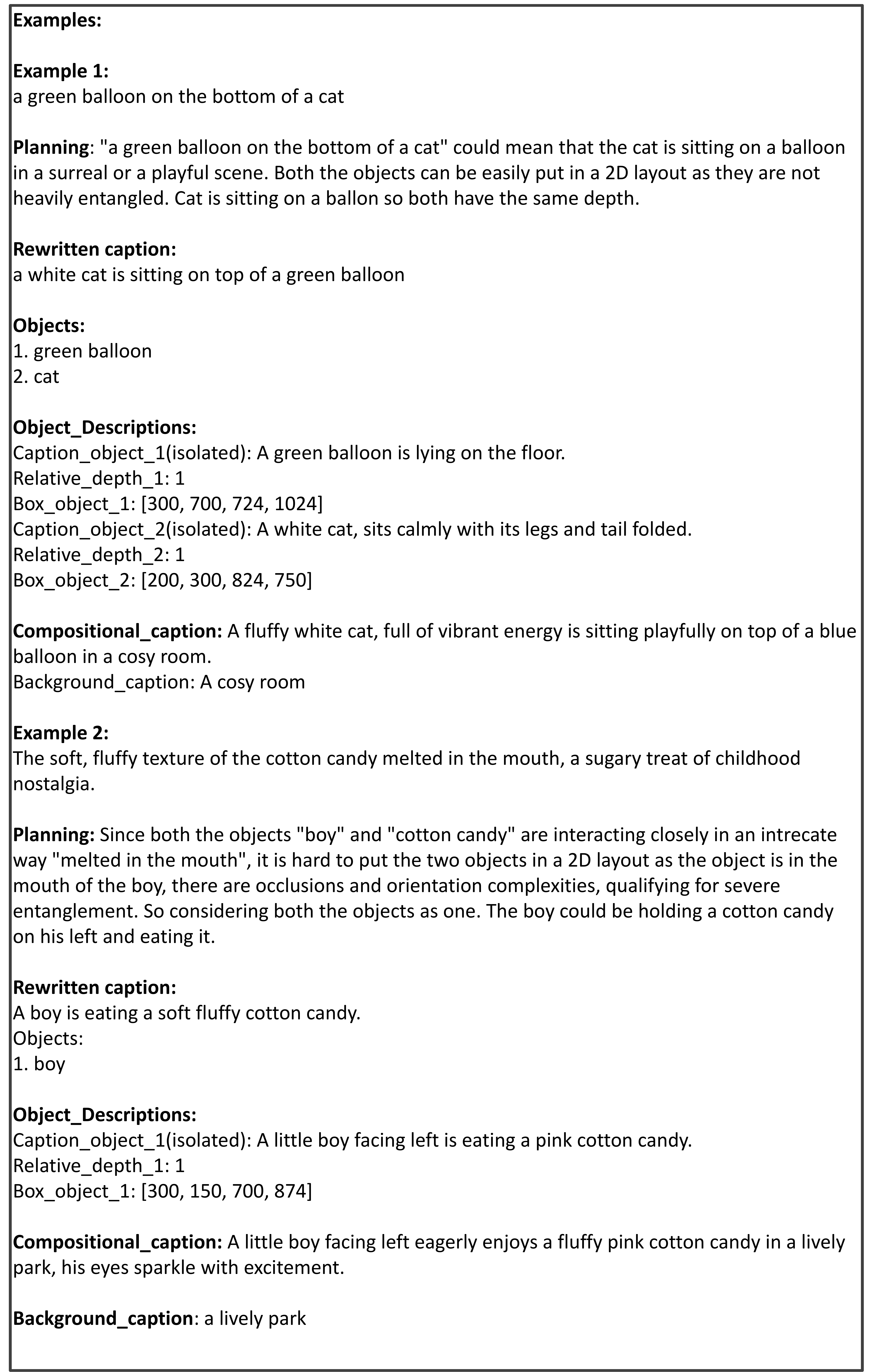}
  \caption{Instructions for LLM planning (continued in the next figure).}
  \label{fig:llm_instruct_3}
\end{figure*}

\begin{figure*}[p]
  \centering
  \includegraphics[height=0.82\textheight]{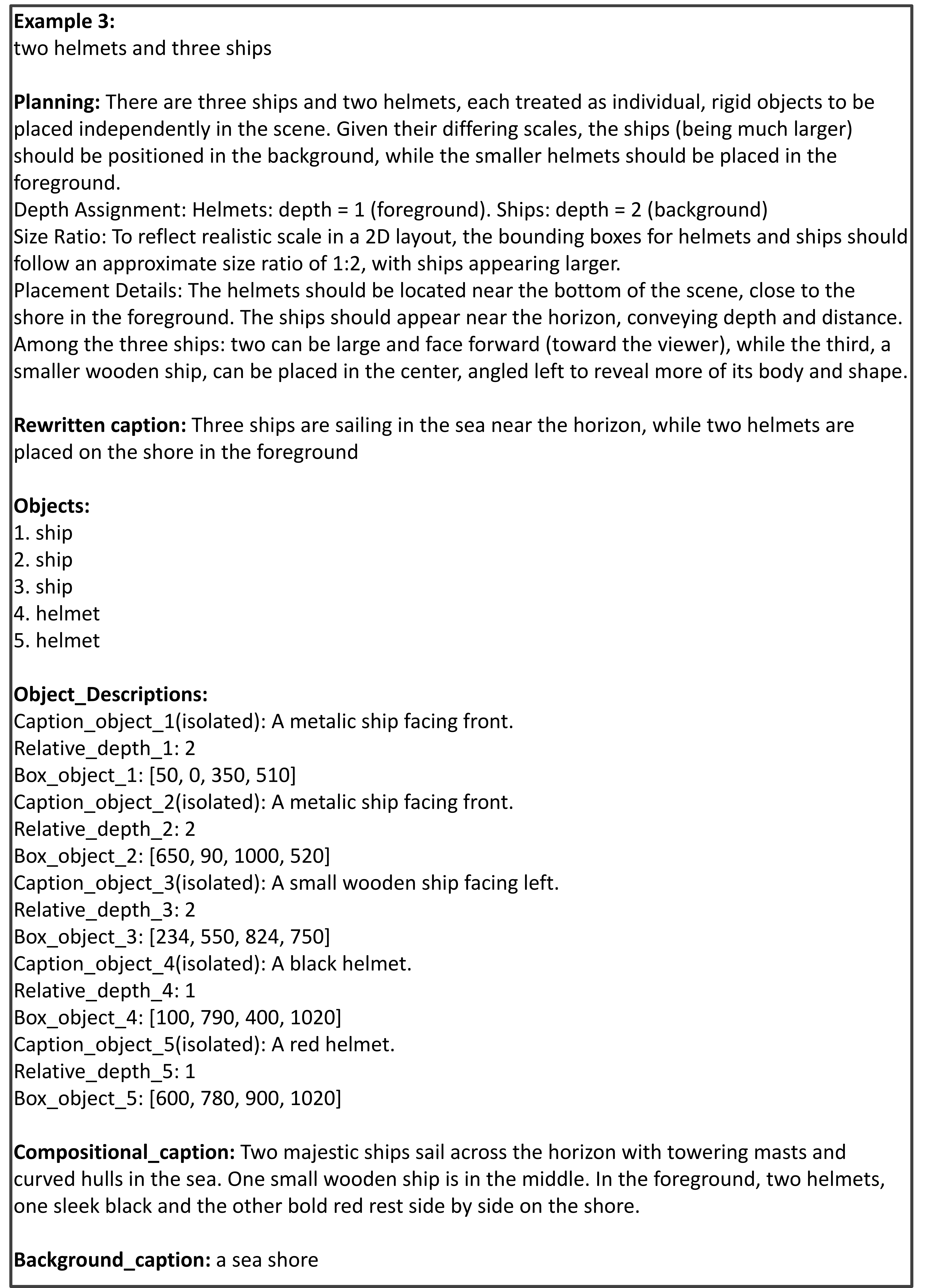}
  \caption{Instructions for LLM planning (continued in the next figure).}
  \label{fig:llm_instruct_4}
\end{figure*}

\begin{figure*}[p]
  \centering
  \includegraphics[height=0.82\textheight]{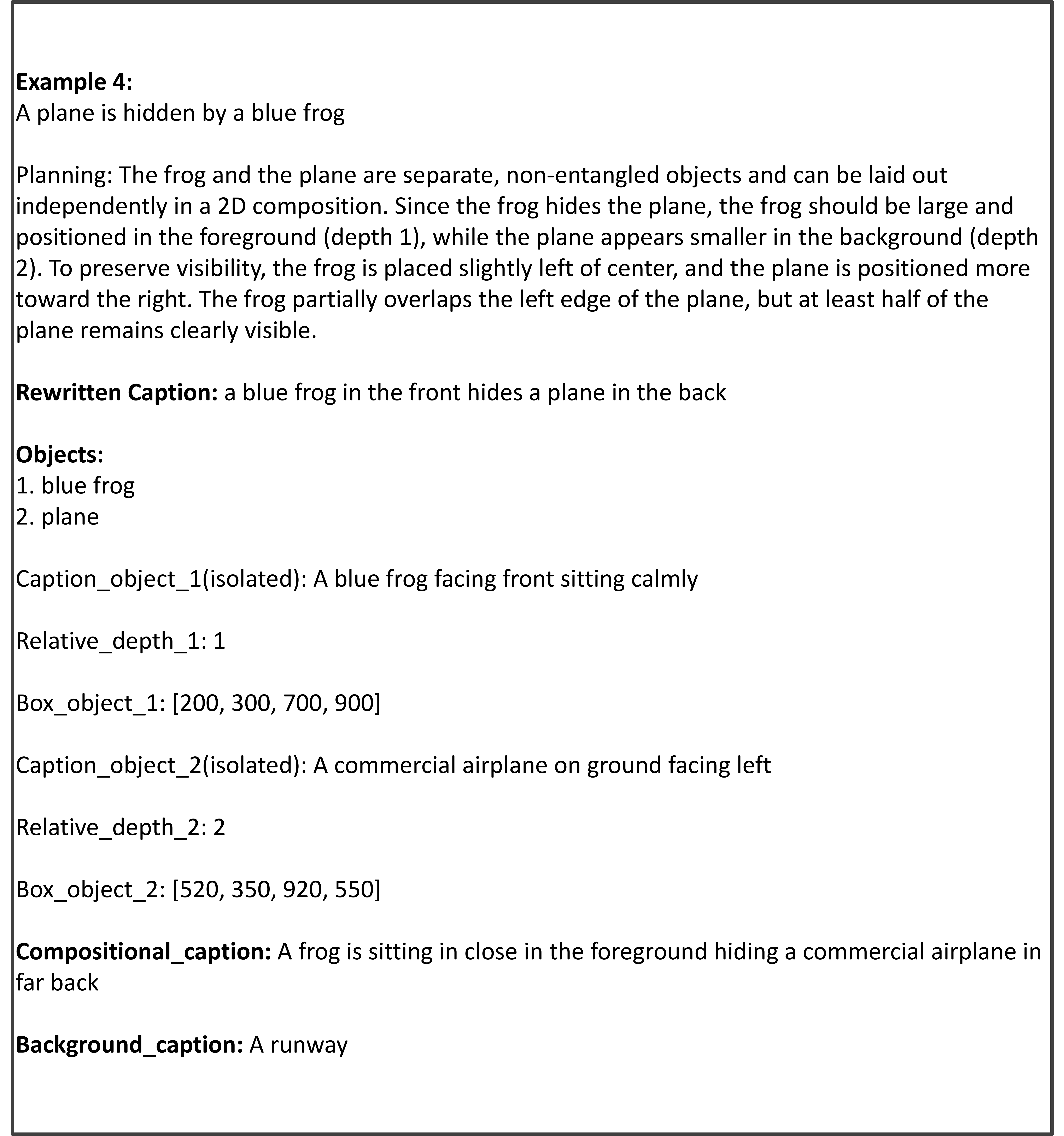}
  \caption{Instructions for LLM planning.}
  \label{fig:llm_instruct_5}
\end{figure*}

\begin{figure*}[p]
  \centering
  \includegraphics[height=0.82\textheight]{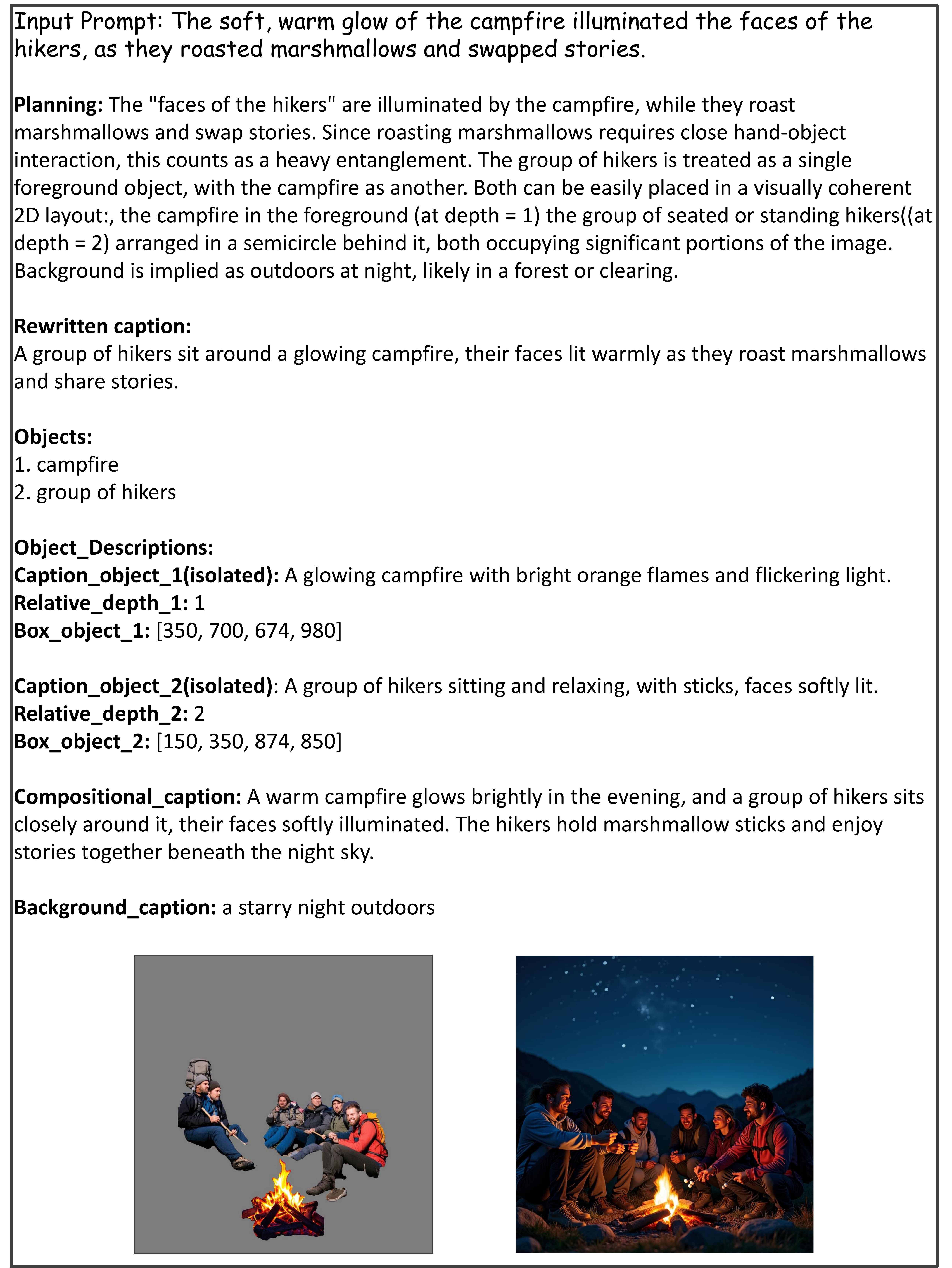}
  \caption{LLM planning for object-prior generation, with the final generated image.}
  \label{fig:llm_prior_examples}
\end{figure*}

Figures~\ref{fig:llm_instruct_1}-\ref{fig:llm_instruct_5} show the detailed instructions for LLM planning.
Figure~\ref{fig:llm_prior_examples} shows the output from the LLM for an example alongside the prior and final image.

\section{3D Metric Evaluation with LLM}
\label{sec:supp_3d_llm_metric}

\begin{figure*}[p]
  \centering
  \includegraphics[width=\textwidth]{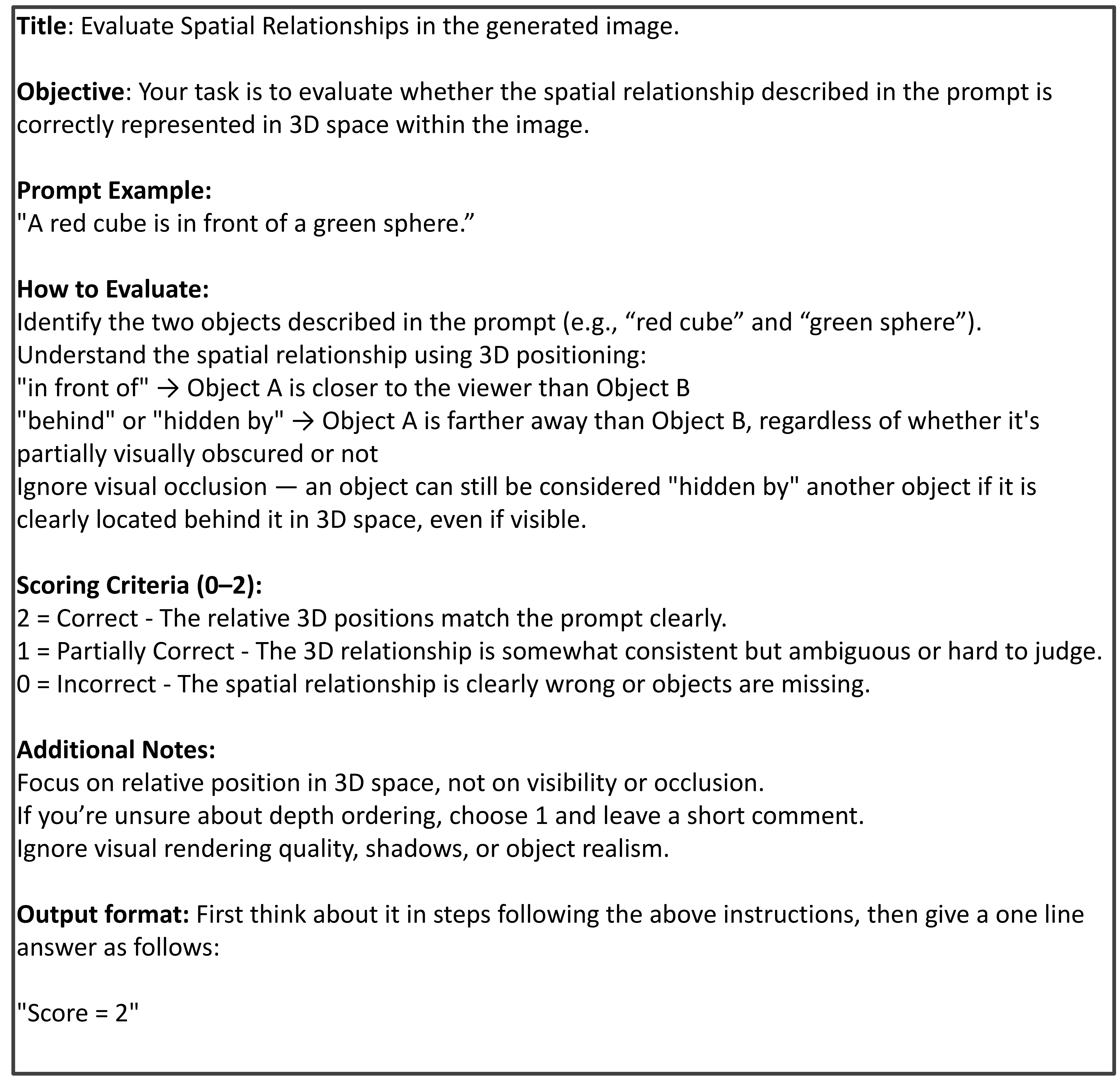}
  \caption{LLM instructions for evaluating 3D-Spatial relations.}
  \label{fig:3D-metric-eval}
\end{figure*}

For the 3D-Spatial category, the original T2I-CompBench metric relies on depth estimation and bounding-box detection, which we found inaccurate and overly punitive to generated images.
Evidence of these limitations appears in Table~10 of the T2I-CompBench++ paper~\cite{huang2023t2icompbench}, where methods showing clear improvements in human evaluation achieve marginal or negative scores with the original metric.
For example, Attn-Exct+SDv2 vs. SDv2 shows a +3.6 improvement in human evaluation but -0.08 in the original metric.
To address this, we introduce an MLLM-based metric using GPT-4.1~\cite{GPT-4o}.
The model is prompted to first identify all required objects and then assess their 3D spatial relations.
Scores are assigned as follows: 0 if objects are missing or 3D relations are wrong, 1 if all objects are present but the 3D relations are ambiguous, and 2 if everything is correct.
We normalize the total score to a 0-100 scale and average over all examples.
Figure~\ref{fig:3D-metric-eval} presents the detailed instructions given to the LLM for evaluating 3D-Spatial relations.

\clearpage

\clearpage

\clearpage

\clearpage

\clearpage

\clearpage

\clearpage

\clearpage

\end{document}